\PassOptionsToPackage{colorlinks=true,allcolors=blue}{hyperref}
\documentclass[a4paper,fleqn]{cas-dc}

\usepackage[numbers]{natbib}

\usepackage{amsmath}
\usepackage{amssymb}
\usepackage{amsfonts}
\usepackage{bm}
\usepackage{graphicx}
\usepackage{float}
\usepackage{subcaption}
\usepackage{adjustbox}
\usepackage{pdflscape}
\usepackage[ampersand]{easylist}
\usepackage[T1]{fontenc}

\usepackage[section]{placeins}

\usepackage[colorlinks=true, allcolors=blue]{hyperref}
\usepackage{tikz}
\usetikzlibrary{
    positioning,
    arrows.meta,
    shapes.geometric,
    shapes.misc,
    fit,
    backgrounds,
    calc,
    decorations.pathreplacing,
    decorations.markings
}

\tikzset{
    >={Stealth[length=2.2mm,width=1.8mm]},
    font=\sffamily\small,
    input/.style    = {rectangle, rounded corners=2pt, draw=blue!55!black,
                       fill=blue!12, minimum width=1.6cm, minimum height=0.85cm,
                       align=center, thick},
    coords/.style   = {rectangle, rounded corners=2pt, draw=olive!70!black,
                       fill=yellow!25, minimum width=1.6cm, minimum height=0.7cm,
                       align=center, thick},
    cell/.style     = {rectangle, rounded corners=3pt, draw=red!65!black,
                       fill=red!12, minimum width=1.7cm, minimum height=1.6cm,
                       align=center, very thick},
    deccell/.style  = {rectangle, rounded corners=3pt, draw=orange!75!black,
                       fill=orange!15, minimum width=1.7cm, minimum height=1.6cm,
                       align=center, very thick},
    output/.style   = {rectangle, rounded corners=2pt, draw=green!45!black,
                       fill=green!18, minimum width=1.6cm, minimum height=0.85cm,
                       align=center, thick},
    attnout/.style  = {rectangle, rounded corners=2pt, draw=violet!70,
                       fill=violet!12, minimum width=1.6cm, minimum height=0.7cm,
                       align=center},
    op/.style       = {circle, draw=black, fill=white, minimum size=0.55cm,
                       inner sep=0pt, font=\small},
    gate/.style     = {rectangle, draw=black!70, fill=gray!10, rounded corners=2pt,
                       minimum width=0.9cm, minimum height=0.55cm, font=\footnotesize},
    block/.style    = {rectangle, draw=black!70, fill=gray!8, rounded corners=2pt,
                       minimum width=1.5cm, minimum height=0.6cm, align=center,
                       font=\footnotesize},
    note/.style     = {font=\scriptsize\itshape, text=black!60},
    flow/.style     = {->, thick},
    state/.style    = {->, very thick, blue!65!black},
    feedback/.style = {->, thick, dashed, red!65!black}
}

\definecolor{Cb}{HTML}{D2E4FA}
\definecolor{Co}{HTML}{FDDBA0}
\definecolor{Cy}{HTML}{FDFAC0}
\definecolor{Cp}{HTML}{E8D8FA}
\definecolor{Cg}{HTML}{C8F0D6}
\definecolor{Cenc}  {HTML}{D2E4FA}
\definecolor{Cdec}  {HTML}{C8F0D6}
\definecolor{CencBg}{HTML}{EDF4FD}
\definecolor{CdecBg}{HTML}{EAF7EE}
\definecolor{Cproj} {HTML}{E8D8FA}
\definecolor{Cattn} {HTML}{FDE9B3}

\newcommand{\U}{u}
\newcommand{\V}{v}

\newcommand{\Mask}{m_{i}}

\newcommand{\Miobs}{M_{i,obs}}

\newcommand{\Nparam}{N}
\newcommand{\Tparam}{T}

\newcommand{\Width}{W}
\newcommand{\Height}{H}

\newcommand{\windspeed}{\sqrt{\U^2 + \V^2}}
\newcommand{\deltatime}{\Delta_{t+\alpha}}

\begin{document}
\let\WriteBookmarks\relax
\def\floatpagepagefraction{1}
\def\textpagefraction{.001}

\shorttitle{Skillful forecasting of offshore winds from satellite scatterometer constellations}
\shortauthors{F.~Pinto et~al.}

\title[mode = title]{Skillful forecasting of offshore winds from satellite scatterometer constellations}

\author[delft,bern]{Francesco Pinto}
\cormark[1]
\ead{f.pinto@tudelft.nl}   

\author[bern]{Luca Lanzilao}

\author[delft]{Paco Lopez-Dekker}

\author[delft,bern]{Angela Meyer}

\affiliation[delft]{organization={Department of Geoscience and Remote Sensing, Delft University of Technology},
    addressline={Stevinweg 1},
    postcode={2628 CN},
    city={Delft},
    country={The Netherlands}}

\affiliation[bern]{organization={School of Engineering and Computer Science, Bern University of Applied Sciences},
    addressline={Quellgasse 21},
    postcode={2501},
    city={Biel},
    country={Switzerland}}

\cortext[1]{Corresponding author}

\begin{abstract}
Accurate intraday forecasts of offshore wind are becoming increasingly important for power system operation and the integration of growing shares of offshore wind energy. Operational forecasts rely predominantly on numerical weather prediction (NWP), which is not optimized for lead times of minutes to hours, where initial-condition accuracy dominates forecast skill. Although satellite scatterometer observations are routinely assimilated into NWP, they have not previously been used directly for forecasting. Here we present WindCastNet, the first satellite-based nowcasting framework for offshore wind speed and direction, introducing a new paradigm for intraday forecasting that learns directly from spatiotemporally irregular satellite observations. WindCastNet predicts offshore wind fields from observations acquired by satellite scatterometer constellations. WindCastNet employs a partial convolutional long short-term memory network that exploits  microwave radar observations from the European MetOp, Chinese HY-2, and Indian Oceansat-3 satellites despite their irregular spatial coverage, asynchronous sampling, and variable revisit times. Spatial observation masks and inter-observation intervals are encoded, while a continuous temporal representation enables forecasts at arbitrary lead times. Evaluated over the North Sea, WindCastNet reduces the root-mean-square error by 23\% and 7\% relative to the HARMONIE-AROME MEPS model at lead times of 1 and 2 h, respectively, and outperforms persistence by 9-15\% during the first three forecast hours. Forecast skill decreases under strong-wind conditions and spatially non-uniform flow.
These results demonstrate that satellite scatterometer constellations can provide an independent and competitive source of short-term offshore wind forecasts, opening new opportunities for renewable energy forecasting but also broader marine weather applications, including tropical cyclone nowcasting.

\end{abstract}

\begin{highlights}
\item We introduce a new paradigm for satellite-based offshore wind forecasting
\item WindCastNet forecasts near-surface winds directly from scatterometer constellations
\item Reduces 1- and 2-hour forecast errors by 23\% and 7\% over state-of-the-art models
\item Outperforms state-of-the-art weather forecasts at lead times up to 2.5 hours
\item First nowcasting from irregular satellite data via spatiotemporal encoding
\end{highlights}


\maketitle
\pagestyle{plain}
\thispagestyle{plain}

\section{Introduction}

Several thousand wind turbines are currently operating in European seas, providing approximately 39 GW of installed capacity \citep{CHIROSCA2022162, remotesensingunveils, windeurope2026}. This capacity is expected to increase almost tenfold in the coming decades, reaching 150-300\,GW by 2050 \citep{denmarkpowerplant}. 

The planned expansion of offshore wind will accelerate the decarbonisation of European energy systems by substantially increasing the share of renewable electricity in the power mix.
However, higher shares of wind energy also pose challenges for power system operation 
because the inherent variability of wind makes balancing electricity supply and demand increasingly complex. Consequently, accurate wind forecasts at lead times of minutes to hours are critical to facilitate wind farm operation, grid integration, electricity trading, energy storage, ancillary service provision, and advanced turbine control.
Improving short-term offshore wind forecasting therefore represents an economically valuable innovation that enhances grid reliability while reducing energy production and system operating costs \citep{mahoney, DUPRE20202373}. 

To overcome the lack of intraday forecasts for spatially extended wind fields, we introduce the first satellite-based wind nowcast framework for near-surface wind fields. We present a novel machine learning approach to nowcast near-surface wind fields using only satellite measurements as input and producing 2D wind field forecasts for lead times up to $3$ hours as output. Our study demonstrates the potential of satellite observations for short-term forecasting of offshore wind fields for minutes to hours ahead. We introduce a wind nowcasting framework for near-surface offshore wind fields. State-of-the-art wind nowcasting methods involve only in-situ and ground-based remote-sensing techniques, thus enabling only pointwise and local-scale (< 10 km) wind monitoring and forecasting. In contrast, the nowcasting of spatially extended wind fields and use of satellite observations and deep neural networks to this end have never been investigated, to our knowledge. We develop and characterise, for the first time, a forecast framework for spatially extended satellite-based wind field estimates and benchmark it with wind forecasts from the state-of-the-art HARMONIE-AROME numerical weather prediction (NWP) model across the North Sea, Europe's largest offshore wind energy region. Specifically, we use the MetCoOp Ensemble Prediction System (MEPS), the operational regional NWP configuration of HARMONIE-AROME employed by the Norwegian Meteorological Institute MET Norway. We leverage scatterometer measurements from satellites in lower Earth orbits. As the scatterometers are not geostationary, their measurements are infrequent and with irregular coverage in space and time. We present a forecast framework able to overcome these challenges.

The main contributions of our study are as follows:
\begin{itemize}
     
\item We introduce a new observation-driven paradigm for offshore wind nowcasting that predicts wind fields directly from satellite observations rather than from numerical weather prediction.

\item We develop \emph{WindCastNet}, the first deep learning nowcasting model designed to learn from spatiotemporally irregular low-Earth-orbit satellite observations.

\item WindCastNet explicitly accounts for irregular spatial coverage, asynchronous observations, and variable satellite revisit times through spatiotemporal encoding.

\item WindCastNet generates continuous-horizon forecasts, enabling predictions at arbitrary lead times without autoregressive inference.

\item WindCastNet outperforms the state-of-the-art forecast system HARMONIE-AROME MEPS  at lead times of up to 2.5 hours.
     
\end{itemize}

This paper is organized as follows. Section~\ref{sec:review} reviews existing methods for offshore wind forecasting. Section~\ref{sec:wind_datasets} describes the datasets used in this study. Section~\ref{sec:model} presents the WindCastNet architecture. Section~\ref{sec:benchmark} introduces the benchmark models, and Section~\ref{sec:performances} presents and discusses the experimental results. Finally, Section~\ref{sec:conclusions} presents conclusions.

\section{Review of wind forecasting methods}~\label{sec:review}
Wind energy is an inherently variable weather-driven energy source. As the shares of wind energy continue to grow, balancing electricity supply and demand becomes increasingly challenging because of the stochastic nature of wind resources.
Accurate forecasts at lead times of minutes to hours (intra-hour to intraday) are therefore becoming essential for the reliable integration of offshore wind into electricity systems. They support wind farm operation, maintenance planning, energy trading, ancillary service provision, grid balancing, and advanced turbine control strategies~\citep{TOSATTO2022, pavingtheway}. Improving short-term wind forecasts is consequently an economically valuable innovation that enhances grid stability while reducing energy production and system operating costs~\citep{DUPRE20202373}. Moreover, because wind turbine power production scales approximately with the cube of wind speed, an error of only 1\% in wind speed can translate into an error of about 3\% in predicted power generation \cite{LYDIA2014452}.

Operational offshore wind forecasts are currently derived predominantly from numerical weather prediction models. NWP provides valuable forecasts from several hours to days ahead but is not optimized for lead times of minutes to a few hours, when forecast skill depends strongly on the accuracy and recency of the initial atmospheric state. Its usefulness at these timescales is constrained by both update frequency and latency. Update frequency describes how often a new forecast cycle is issued, whereas latency is the time between an observation being acquired and a forecast incorporating that observation becoming available to users. In NWP systems, observations must first be assimilated within an analysis cycle, followed by numerical model integration and post-processing. Consequently, even observations included in the most recent analysis may only influence forecasts available to users substantially later.

Regional NWP systems are typically re-initialised every few hours. For example, the HARMONIE-AROME MEPS variant, used operationally by several national meteorological services surrounding the North Sea and Baltic Sea, provides forecast updates every three hours~\citep{KNMI2024, TheOperationalForecastProcessatMetCoOp}. Statistical and machine-learning models are also computationally substantially less demanding than NWP models, allowing forecasts to be generated and updated rapidly after new observations become available. The combination of multi-hourly update cycles, computational costs, and processing latency means that NWP tends to delay the incorporation of newly acquired observations by at least several hours, limiting their ability to support operational decisions under rapidly evolving wind conditions. 

Nowcasting addresses these gaps by producing high-accuracy forecasts from minutes to a few hours ahead \cite{nowcasting}. Unlike NWP, observation-driven nowcasting systems can process newly available measurements without waiting for a full data-assimilation and numerical-integration cycle, requiring substantially less computation than NWP. Nowcasting can therefore offer distinct operational advantages: higher forecast accuracy as a result of more frequent forecast updates and lower latency between observation acquisition and forecast delivery, in addition to lower computational cost. Although dynamical forecast models remain indispensable at longer lead times, nowcasting can complement them by providing forecasts that incorporate recent observations more rapidly and are refreshed more frequently during the first few forecast hours.

Despite this potential, wind nowcasting remains largely underexplored relative to day-ahead and longer-range wind forecasting~\citep{asurveyonwindpower, Tawn}. 
Existing wind nowcasting methods rely on local in-situ observations, including cup anemometers, meteorological towers, and wind lidars located near wind farms~\citep{HU201917, LIU2021573, minutescaledetection}. Most approaches formulate nowcasting as a time-series forecasting problem using statistical models such as ARIMA or machine-learning architectures such as long short-term memory networks~\citep{asurveyonwindpower, mlapproachesforwindpowerforecasting}. These methods have demonstrated promising performance for localised forecasting based on in-situ wind measurements, lidar observations, or site-specific downscaling~\citep{DUPRE20202373, winddirectionforecasting, HU201917}. However, their spatial coverage is typically limited to at most a few kilometres around the measurement location. They therefore contain little information about upstream wind conditions and cannot adequately anticipate approaching mesoscale changes, including frontal passages and other spatially evolving wind structures. Consequently, they are not suitable for regional offshore wind nowcasting across extensive domains such as the North Sea or Baltic Sea. In-situ and remote-sensing techniques for monitoring offshore wind resources have been reviewed by~\citep{offshorewindsmappedfromsatellite, remotesensingobservation, reviewmethodologiesoffshore}.

While localized nowcast methods of wind speed and direction have been proposed, observation-based regional offshore wind field nowcasts are lacking. 
Satellite observations enable monitoring over large spatial domains. However, their potential for spatially extended wind field nowcasting has remained unexplored. 
Several studies have investigated satellite-based radar measurements for monitoring wake effects in offshore wind farms \citep{HASAGER2026115369, highresolutionoffshore, haeser2}, but not for wind field nowcasting. This gap contrasts sharply with the growing demand for accurate regional wind forecasts at intra-hour and intraday lead times across a wide range of marine applications, including offshore wind energy and wind-farm operations, but also tropical cyclone monitoring and forecasting.

\section{Satellite-derived wind observations}
\label{sec:wind_datasets}
Satellite-based scatterometers provide global estimates of near-surface wind fields over oceans, where wind observations are sparse. By providing continuous day-and-night wind estimates over the oceans at all weather conditions, scatterometers can significantly improve the representation of near-surface winds in weather models. Scatterometer observations are operationally assimilated in numerical weather prediction systems by weather forecast centres, such as the European Centre for Medium-Range Weather Forecasts (ECMWF), which has been shown to improve atmospheric analyses and significantly reduce surface wind root mean square errors \citep{ascathersbach, AssessmentofAssimilatingASCATSurfaceWindRetrievalsintheNCEPGlobalDataAssimilationSystem, ImpactofScatterometerSurfaceWindDataintheECMWFCoupledAssimilationSystem}. 
At forecast lead times of minutes to a few hours, forecast errors are strongly affected by uncertainties in the initial atmospheric state. Despite their success in numerical weather forecasting, the potential of scatterometers for direct satellite-based forecasting at these lead times remains unexplored.

A scatterometer is a satellite-borne microwave radar that retrieves near-surface wind fields over oceans from the radar backscatter of the sea surface. Scatterometers estimate near-surface wind speed and direction indirectly by measuring the wind-induced changes in ocean surface roughness. Scatterometers emit microwave pulses towards the ocean and measure the fraction of the signal scattered back to the satellite. Unlike optical sensors, they are largely unaffected by cloud cover and operate independently of solar illumination. Their microwave pulses are backscattered from centimetre-scale waves generated by the local wind. Stronger wind increases the roughness of the sea surface and, as a result, the magnitude of the backscattered microwave signal. The backscatter intensity also depends on how the scattering surface waves are oriented relative to the scatterometer's viewing direction. To be able to infer the wind direction, scatterometers observe the same ocean surface from multiple viewing angles during each overpass.

In this study, we leverage scatterometer observations to develop a framework for satellite-based intraday forecasts of near-surface wind speed and direction. Our forecasts rely on scatterometer measurements from a constellation of six polar-orbiting satellites: Two Advanced Scatterometer (ASCAT) \citep{ASCAT, ASCAT-B, ASCAT-C} instruments onboard the MetOp- B and -C satellites; three HY-2 Scatterometer (HSCAT) \citep{HSCAT-B, HSCAT-C, HSCAT-D} instruments onboad the HY-2B, -2C and -2D satellites, and the Oceansat-3 Scatterometer (OSCAT) \citep{SANKHALA20251945, OSCAT} onboard the Oceansat-3 satellite, as shown in Table \ref{table:satellite_table}. Together, this constellation provides approximately eight to nineteen satellite overpasses per day over the North Sea since November 2021.

ASCAT is a microwave scatterometer on MetOp operating in the C-band at approximately 5.3 GHz, corresponding to 5.6 cm microwaves, whereas the HSCAT and OSCAT instruments onboard the HY-2B/C/D and Oceansat-3 satellites are Ku-band scatterometers emitting at 13.5 GHz, corresponding to wavelengths of approximately 2.2 cm (Table \ref{table:satellite_table}). While the C-band is less sensitive to rain, Ku-band instruments enable higher sensitivity to small-scale sea surface roughness \citep{10281803}. ASCAT employs a fixed fan-beam antenna that provides a fixed wide radar beam, whereas the HSCAT and OSCAT instruments use rotating pencil-beam antennas that scan narrow microwave radar beams across the ocean surface from multiple viewing angles. The operational ASCAT Level-2 product used here \cite{ASCAT-C} is delivered on a 12.5\,km Wind Vector Cell grid, then upscaled to 25\,km horizontal resolution in order to fit the other satellite observations, while the other L2 products, described below, are already provided at a 25\,km resolution.

MetOp satellites fly in a sun-synchronous orbit at approximately
817 km altitude, with a 09:30 local-time
equator crossing, passing over the North
Sea occur at approximately the same local time each day.
HY-2B is in a sun-synchronous orbit with a 06:00 local-time equator crossing, whereas HY-2C and HY-2D are in non-sun-synchronous orbits at 66 degree inclination,
so that their overpass times over the North Sea drift
through the day, providing temporally diversified sampling.
A 25\,km HSCAT L2 wind vector product \cite{verhoef2024hscatpum, HSCAT-B, HSCAT-C, HSCAT-D} is used in this work: the 25\,km Level-2 wind product from the EUMETSAT Ocean and Sea Ice Satellite Application Facility (OSI SAF) at the Royal Netherlands Meteorological Institute (KNMI). OSCAT \cite{verhoef2024hscatpum, SANKHALA20251945, OSCAT} is in a sun-synchronous orbit with an approximate 12:00 local-time equator crossing. 

All three instruments are processed by KNMI within OSI
SAF,
so that the retrieved winds are inter-calibrated and share consistent bias characteristics
relative to buoy and ECMWF reference winds~\citep{verhoef2024hscatpum}.
Triple-collocation analysis with moored buoys and ECMWF forecasts
indicates a wind-speed bias below $0.01$\,m/s and zonal/meridional wind-component error standard deviations of approximately
$0.5$-$0.8$ m/s for the 25km HSCAT and OSCAT products and
of comparable magnitude for ASCAT, well within the OSI SAF
requirement of $2$ m/s \citep{verhoef2024hscatpum}. These
accuracies are at least an order of magnitude smaller than the
forecast errors reported in Section~4 for both WindCastNet and MEPS,
so that the scatterometer retrievals can be safely treated as
high-quality reference observations for the present study.

\begin{table}[pos=H]
\centering

\scriptsize
\begin{tabular}{lllllll}
\hline
 Instrument & Satellite & Orbit & Alt. & Freq. & Agency  \\
\hline
ASCAT  & MetOp-B & sun. & 830 & 5.255 & EUMETSAT \\
ASCAT  & MetOp-C & sun. & 827 & 5.255 & EUMETSAT \\
HSCAT &  HY-2B & sun. & 973 & 13.256 & NSOAS  \\
HSCAT  &  HY-2C & drift. & 973 & 13.256 & NSOAS \\
HSCAT  &  HY-2D & drift. & 1336 &  13.256 & NSOAS \\
OSCAT  & Oceansat-3 & sun. & 723 & 13.515 &  ISRO \\
\hline
\end{tabular}
\caption{The scatterometer instruments used in this study. The columns 'Alt.' and 'Freq.' provide the satellite altitude in km and the radar frequency in GHz, respectively. 'sun.' and 'drift.' denote sun-synchronous and drifting orbits, respectively.}
\label{table:satellite_table}
\end{table}

\begin{figure*}[pos=H]
    \centering
    \begin{subfigure}{0.48\textwidth}
        \centering
          \includegraphics[width=\linewidth]{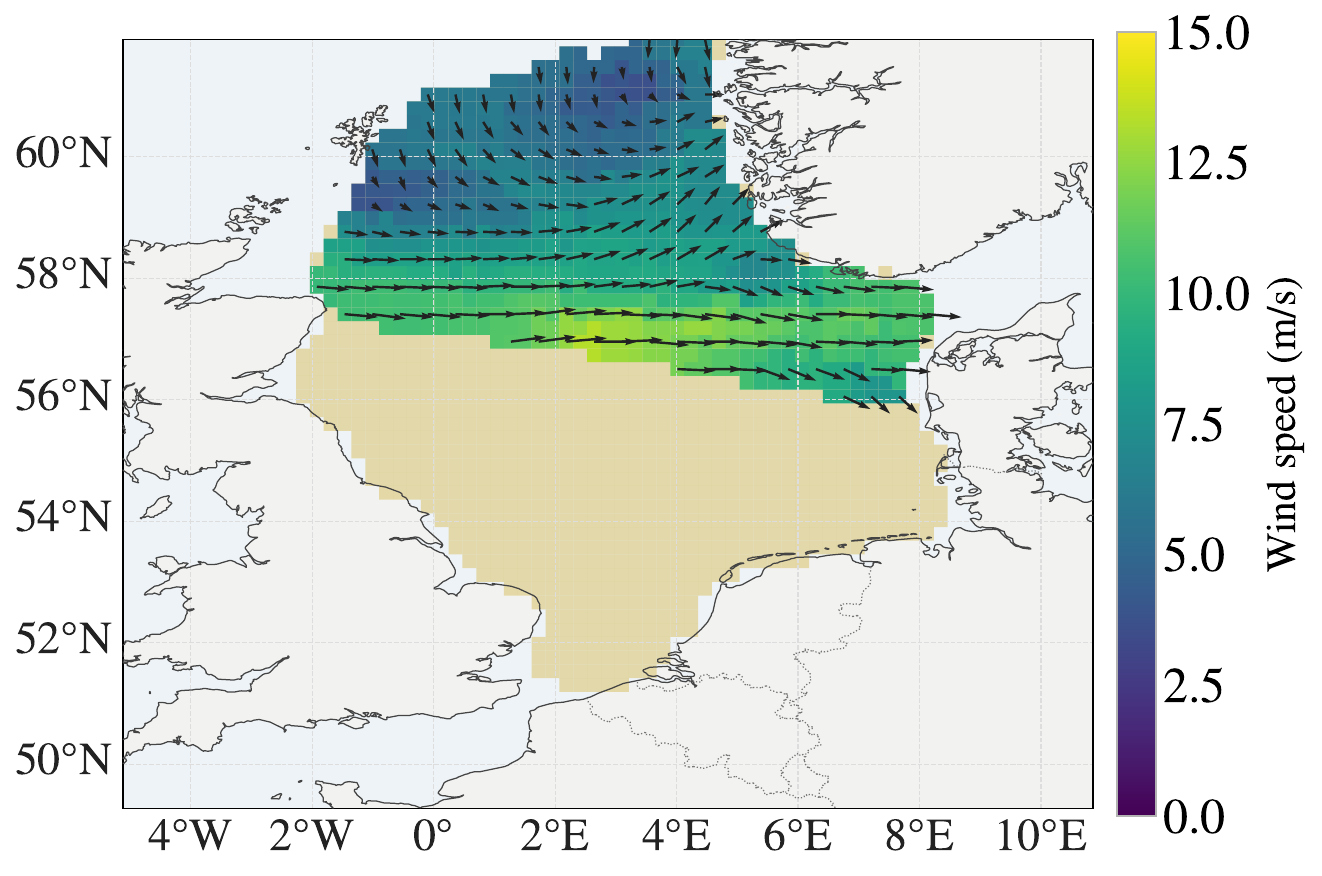}
        \caption{Scatterometer wind measurement on 2023-01-01, 08:09am. }
        \label{fig:scatterometer_measurement_1}
    \end{subfigure}
    \hfill
    \begin{subfigure}{0.48\textwidth}
        \centering
        \includegraphics[width=\linewidth]{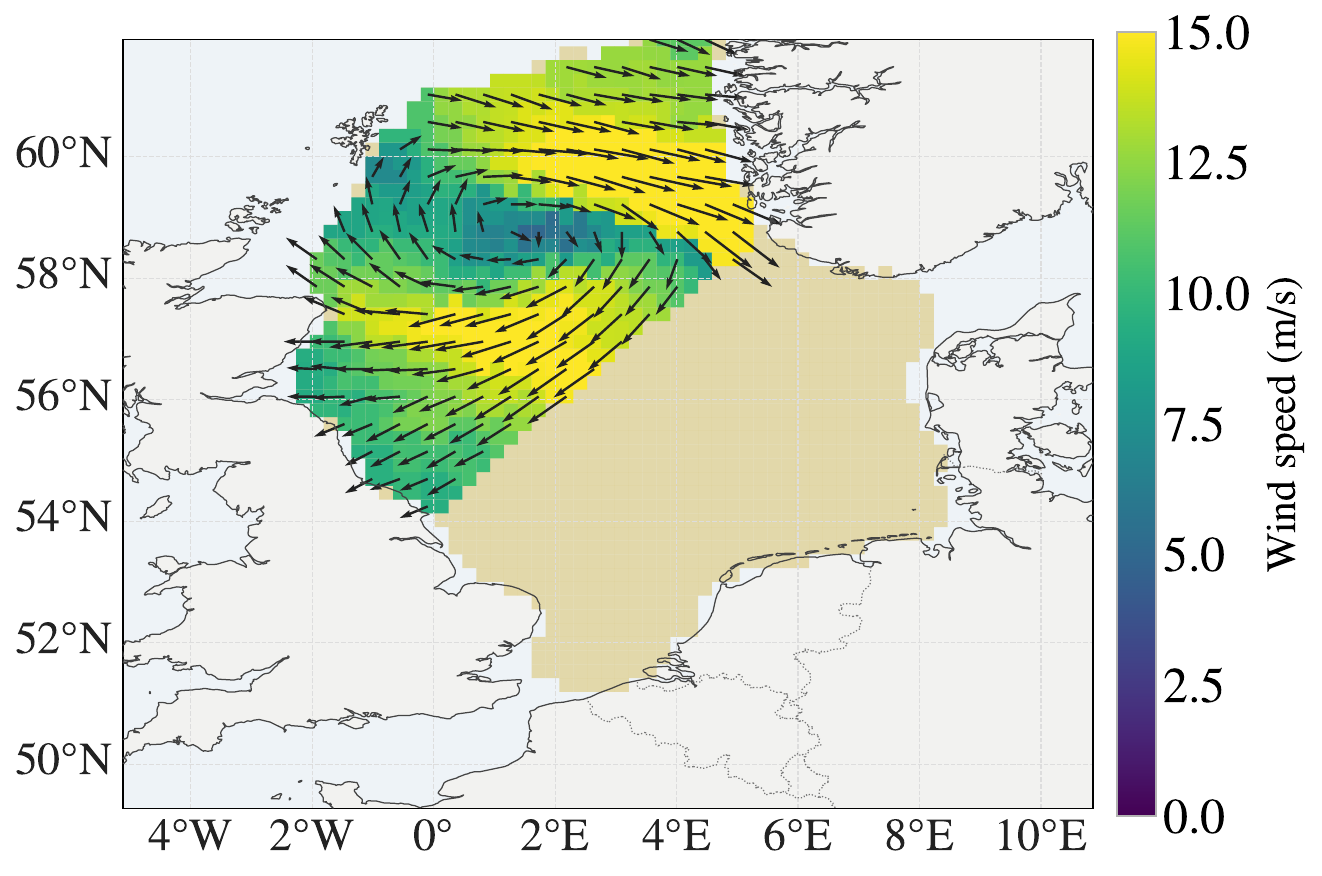}
        \caption{Scatterometer wind measurement on 2023-01-04, 07:57am.}
        \label{fig:scatterometer_measurement_2}
    \end{subfigure}
    \caption{Scatterometer-derived wind fields randomly selected from the training set. The underlying beige shading marks the study domain. All times reported in this study are UTC times.}
    \label{fig:two_images}
\end{figure*}
All the above-mentioned satellites provide near-surface wind estimates at approximately 10 meters height.
Since the coarsest spatial resolution available among the scatterometer products is $25 \times 25$ km, we resample all wind observations to this resolution in order to obtain a consistent input grid. 

Operational 10-m wind forecasts from the HARMONIE-AROME model configuration of MEPS are used as state-of-the-art operational forecast benchmark~\citep{mepslink}, as detailed in section~\ref{sec:benchmark}. The MEPS forecasts are provided at hourly forecast lead times and updated every $3$ hours at 00:00, 03:00, 06:00, 09:00, 12:00, 15:00 and 18:00.
To ensure that the surface wind forecasts from satellite scatterometers and from MEPS are comparable, we upscaled the MEPS forecasts to the same $25 \times 25$ km resolution.
The training, validation and test periods used in this study were the November 2021 to December 2024, January to April 2026 and the full year 2025, respectively, as shown in Fig. \ref{fig:dataset_timeline} and described in more detail in Appendix Section~\ref{sec:train_val_test}. The spatial study domain is the North Sea in Northwestern Europe, as shown in Fig. \ref{fig:scatterometer_measurement_1} and \ref{fig:scatterometer_measurement_2}, where most of Europe's offshore wind power has been operating.

Unlike instruments on geostationary satellites, the scatterometers do not observe the same region continuously but orbit Earth in near-polar orbits approximately every 100 minutes, following different ground tracks at each overpass. 
As a result, they complete roughly 14–16 orbits per day and provide measurements that are irregular in both space and time. Two examples of resulting wind observations over the North Sea are shown in Figs. \ref{fig:scatterometer_measurement_1} and \ref{fig:scatterometer_measurement_2} for the HSCAT instrument for illustration. 
The MetOp ASCAT, HY-2B HSCAT and Oceansat-3 OSCAT scatterometers operate in sun-synchronous orbits, resulting in overpasses over the North Sea that occur at approximately the same local time each day. 
In contrast, the HY-2C and HY-2D satellites are not sun-synchronous and revisit the North Sea at varying times, providing a more temporally distributed sampling of the wind field. Due to the large swath width of the HSCAT instruments of approximately $1300$ km, consecutive overpasses over the North Sea can occur within short time intervals. A visual illustration of how the resulting overpass frequencies are distributed by time of day and by satellite is shown in Fig.~\ref{fig:dataset_overpasses}.

Due to variations in satellite orbits, the wind dataset used in this study contains training, validation, and test samples that either fully cover the North Sea (Fig. \ref{fig:scatterometer_measurement_1}) or only partially cover it (Fig. \ref{fig:scatterometer_measurement_2}). Samples containing fewer than 200 valid data points over the study domain were considered as unsuitable as model inputs and therefore discarded. Since a complete wind field contains approximately $1400$ data points, we aimed for a spatial context that contains at least approximately $15\%$ of the possible measurements. Using significantly fewer than 200 data points would lead to a lack of context, while requiring a significantly higher number would entail a lack of samples. 
After removing samples with incomplete coverage, the dataset contains a total of $19630$ usable samples from November 2021 to April 2026, of which we actually use $13223$ as a result of balancing the recency of observations (time window length) against the minimum number of observations per time window, as described in Appendix Section~\ref{sec:dataset_tradeoff}.

\section{Model architecture}
\label{sec:model}

Our model, WindCastNet, predicts the next frame of the near-surface wind field from a sequence of
incomplete observations. Each input frame carries an availability mask that marks which grid cells are observed, and the backbone is a partial convolutional long short-term memory network (PConvLSTM) \citep{pconvlstm} that combines partial convolutions \citep{partialconvolutions} with a
convolutional long short-term memory (LSTM) network \citep{convlstm}. We additionally introduce a lead-time
($\deltatime$) conditioned output head that lets a single trained model generate forecasts at a requested horizon: we use a combination of time embedding with Time2Vec \citep{time2vec} and a conditioned layer with Feature-wise Linear Modulation (FiLM) \citep{film}. FiLM is a conditioning method that allows neural networks to adjust their predictions based on auxiliary input by shifting feature maps. In our study, FiLM utilises the forecast lead time to modulate the output head in order to allow a single model, WindCastNet, to forecast at arbitrary lead times without using separate output heads.

\begin{figure*}[pos=H]
    \centering
    \includegraphics[width=1\linewidth]{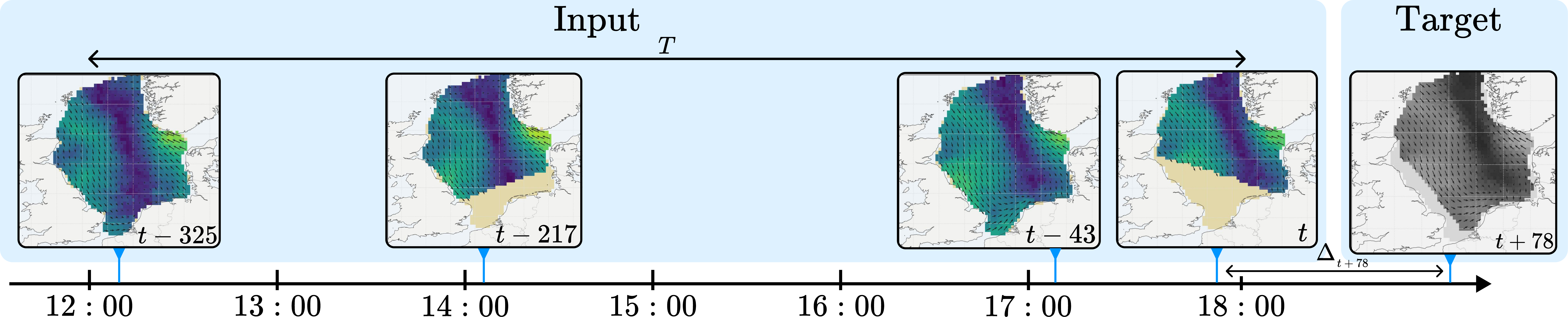}
    \caption{Illustration of the input data extraction process. In this example, we select $N=4$ samples within a time window of $T=6$ hours. The input sequence of the wind measurements is irregular in time and space: the first $N-1$ samples are extracted $325$, $217$ and $43$ minutes before the last input timestamp at time $t$. The forecast lead time is $78$ minutes.}
    \label{fig:input_data}
\end{figure*}

\subsection{Model design for irregular spatiotemporal coverage}
\label{sec:spatiotemporal_irregularity}
The spatial and temporal irregularity of the wind fields poses a considerable challenge for the development of a machine learning model that aims to predict a spatially complete wind field while maintaining temporal regularity. 
The model design therefore needs to account for both the varying spatial coverage of the study domain by the wind field samples and their irregular acquisition times.
After extracting the wind measurements collected over the study domain, we gather the zonal and meridional wind components $u$ and $v$ into input sequences controlled by two hyperparameters: a temporal window $\Tparam \in \mathbb{N}^+$, expressed in hours, that defines the time interval from which the wind measurements are drawn, and the number of wind measurements $\Nparam \in \mathbb{N}^+$ sampled within that interval. An illustration of the input sample extraction from a time window of length $\Tparam$ is shown in Fig.~\ref{fig:input_data}.
The resulting input tensor has dimensionality $(\Nparam, 2, \Width, \Height)$, where $2$ represents the wind components ($u$ and $v$), and $\Width$ and $\Height$ represent the spatial width and height of the grid. 
Since each grid cell corresponds to an area of $25 \times 25$ km, and the study domain covers approximately 1775\,km $\times 1425$\,km, the spatial dimensions become \Width =  71 and \Height =  57.

This formulation introduces a trade-off in the choice of $\Nparam$ and $\Tparam$. 
A long time window $\Tparam$ may include input samples that are too distant in time from the target timestamps, thus providing limited relevant information for the forecast. 
Conversely, a short time window $\Tparam$ may result in an insufficient number of input samples $\Nparam$, since satellite overpasses can provide sparse and irregular samples over several hours. The optimal combination of hyperparameters $(\Tparam,\Nparam)$ is obtained through a grid search over the hyperparameter space shown in Fig.~\ref{fig:tradeoff}.
We refer to Appendix section~\ref{sec:dataset_tradeoff} for further details.

To encode the spatial irregularity of the dataset, we augment each input frame with a binary availability mask $\Mask \in \{0,1\}^{\Width \times \Height}$ associated with the $i$-th wind measurement, which is $1$ at the grid cells where an observation exists and $0$ elsewhere. Rather than being concatenated with the wind components, the mask is passed alongside them to the partial convolutions of the network, which renormalises each output by the number of valid grid cells in its receptive field. This allows the model to distinguish observed from unobserved regions without treating missing values as zeros, and to propagate the validity mask through its internal state.
In our setting, we select an input sequence length of $\Nparam = 6$ within a maximum time window length of $T = 12$ hours, as explained in subsection \ref{sec:dataset_tradeoff}. The input is therefore assembled from the two wind components $u$ and $v,$ forming a tensor of shape $(6, 2, \Width, \Height)$, paired with the availability mask $\Mask$ of shape $(6, 1, \Width, \Height)$.

The dataset exhibits irregular temporal sampling because the satellite overpasses do not occur at fixed time intervals. Therefore, the model must also account for the 
temporal distance between the time $t$ of the most recent observation and the predicted frame. This temporal distance is a scalar forecast lead time $\deltatime \in \mathbb{R}$, expressed in hours and defined as the gap between the
most recent input observation and the requested future target time (the forecast lead time), so that a
lead time of $60$ minutes corresponds to $\Delta_{t+60} = 1$ and a lead time of $30$ minutes to $\Delta_{t+30} = 0.5$. The model does not learn separate outputs for fixed lead times (e.g., +1\,h, +2\,h, +3\,h). Instead, it learns a continuous mapping from lead time to forecast, allowing it to make predictions at any requested forecast horizon. So the desired forecast horizon becomes an input to the model rather than being hard-coded into the network architecture.
Since the model produces a single-step forecast at an arbitrary horizon, $\deltatime$ conditions only the output head, through a Time2Vec embedding and FiLM modulation (Section~\ref{model_structure}), so that a single model can predict at
the exact horizons at which scatterometer overpasses become available
(e.g., $20$, $57$ or $61$ minutes ahead) instead of a fixed temporal grid.

\subsection{Model structure} 
\label{model_structure}
\begin{figure}[pos=H]
    \centering
    \includegraphics[width=6cm]{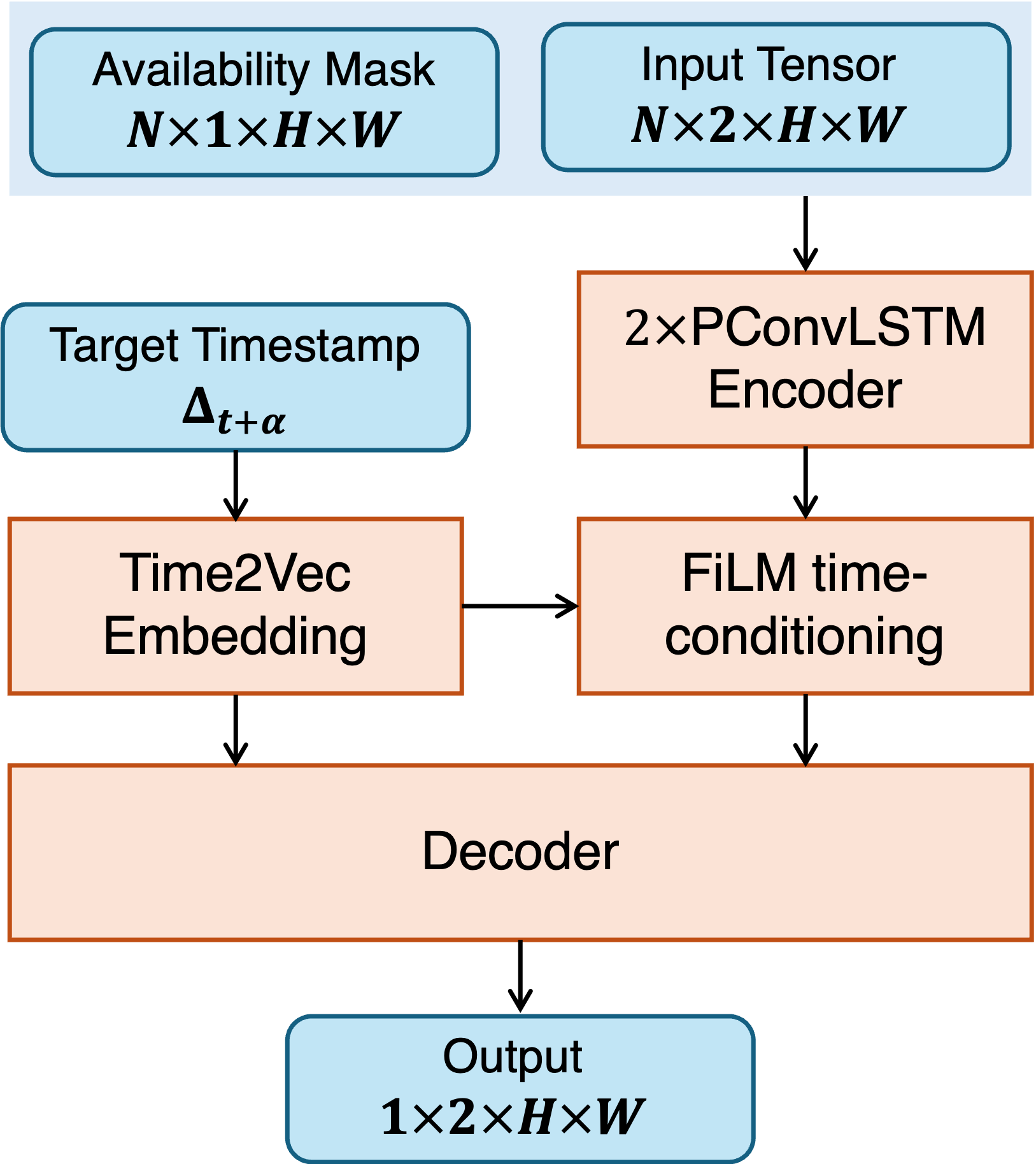}
    \caption{WindCastNet model structure. The model takes three different inputs: the $N$ past wind fields, the availability mask, and the target timestamp that is embedded through the Time2Vec layer and concatenated to the PConvLSTM output to produce the forecast with the FiLM and decoder. The output is a single wind field at the desired forecast lead time.}
    \label{fig:model_structure}
\end{figure}

The model encodes the hidden state of the input sequence with two PConvLSTM cells (Section \ref{sec:partial_convlstm}) that produce a \textit{hidden state}, a \textit{cell state} and a \textit{mask's hidden state} that are recursively re-injected in the cell.
After the input encoding, the time is embedded through a \textit{Time2Vec} vector representation: the hidden state and the vector representation are then used as input for the time-conditioning block that produces a \textit{conditioned hidden state} with a FiLM architecture (Section \ref{sec:leadtimeconditioning}). Finally, the FiLM and Time2Vec output produce the forecast at time $\deltatime$, where $t$ is the last timestamp and $\alpha$ denotes the number of minutes ahead to be predicted. Fig. \ref{fig:model_structure} shows the model structure.

The encoder stacks two PConvLSTM cells.
At each time step, the first layer is driven by the observation
$(x_i, \Mask)$ and every subsequent layer is driven by the hidden state of the layer below together with its propagated
mask; the recurrence of each layer consumes the raw (un-normalised) hidden
state. All states and masks are initialised to zero. After the full
sequence has been encoded, the top-layer hidden state $h^{\mathrm{top}}$
at the final step and its mask $m^{\mathrm{top}}$ are projected to the
predicted next frame by a final partial convolution,
\begin{equation}
\hat{x} = \mathrm{PConv}\big(h^{\mathrm{top}},
m^{\mathrm{top}}\big),
\label{eq:head}
\end{equation}
yielding a forecast whose support is consistent with the propagated mask. The lead-time conditioning is discussed in more detail in Appendix subsection \ref{sec:leadtimeconditioning}.

Through a grid search over the
hidden width, kernel sizes and number of layers, we select an encoder with
hidden dimension $128$, input-to-state and state-to-state kernels of size $7$,
and $2$ stacked layers. The resulting model has approximately $9.9 \times 10^{6}$
parameters.

The model is trained on wind observations from November 2021 to December 2024, validated from January 2026 to April 2026, and evaluated on a test set covering the entire year 2025. This temporal split ensures a strict separation between training and validation, preventing information leakage.
The model evaluation and loss metrics are provided in Appendix subsection \ref{sec:lossmetrics}. To mitigate border artifacts, the loss is computed only over a central spatial patch of the North Sea during validation. During training, the loss is evaluated on the $u$ and $v$ produced by the model, but the performance is assessed using the root mean squared error (RMSE) of wind speed (expressed in $m/s$) between predictions and satellite observations, evaluated only where valid measurements are available.
Due to the irregular spatial and temporal coverage of satellite data, the number of valid grid cells varies across lead times, resulting in different sample sizes for each evaluation horizon.

\section{Benchmark models}~\label{sec:benchmark}
We next introduce the benchmark models against which WindCastNet is evaluated.
\subsection{HARMONIE MEPS}~\label{sec:MEPS}
The MetCoOp Ensemble Prediction System (MEPS) \citep{TheOperationalForecastProcessatMetCoOp} is the convection-permitting limited-area ensemble NWP model from the MetCoOp collaboration of the meteorological agencies of Estonia, Finland, Latvia, Norway, and Sweden. Built on the HARMONIE-AROME reference system \citep{KNMI2024}, 
MEPS forecasts an ensemble of 30 members on a grid of $960 \times 1080$ points at uniform 2.5-km horizontal resolution with 65 hybrid sigma-pressure terrain-following vertical levels extending from the surface to 10 hPa. MEPS is integrated in single precision with a 66-h forecast range and produces hourly output, which is disseminated approximately 2\,h after the analysis time whose observation cutoff is 75 min \cite{TheOperationalForecastProcessatMetCoOp} meaning the time limit after which incoming weather data is locked out and can no longer be included in the current forecast cycle. So when a forecast is distributed to users, the most recent scatterometer observations included in that forecast became available more than 3.25 hours ago. As the observation cutoff time does usually not coincide with a scatterometer overpass time, the latest scatterometer observations will in most cases have been made available more than 4\,h ago.
Each ensemble member is reforcast every 3 h, but production is spread in time so that every hourly cycle updates five members to even the computational load across geographically distributed HPC platforms 
and provide frequent updates.
The unperturbed control member takes its lateral boundaries from ECMWF HRES while the 14 perturbed members each run twice per 3 h drawing on 28 individual members from the ECMWF Integrated Forecasting System (IFS).

For each scatterometer overpass at valid time $t$, we use the most recent MEPS forecast cycle initialised at $t$ or earlier, and extract the +1\,h to +3\,h lead times from that cycle. In an operational setting, however, NWP forecasts are not available instantaneously at their initialisation time: the process of data assimilation, model integration and product dissemination introduces a typical latency of approximately two hours before a given cycle is delivered to end users. 
Because WindCastNet is deterministic, we compare it against the MEPS control member
\citep{mepslink} rather than against the full ensemble. The control member is run in
deterministic mode, i.e.\ as a conventional NWP run comprising data assimilation followed by
numerical integration every 3\,h.

\subsection{Persistence Model}
We adopt persistence as an additional
baseline common in the nowcasting literature \cite{persistence}. Persistence provides an interpretable lower bound that any data-driven model is expected to outperform to be considered skilful.
The persistence baseline assumes that the wind field is invariant between the model initialisation time and the validation time. The persistence forecast at lead time $\tau$ is constructed by pairing two scatterometer observations of the same location separated by $\tau$. 
In our setting, the initialisation time $t$ is defined as the timestamp of
the most recent scatterometer observation in the input sequence
(the $N$-th frame; see Section \ref{sec:spatiotemporal_irregularity}), and the persistence forecast at a
lead time $\tau$ is given by
\begin{equation}
  \hat{\mathrm{v}}_{\mathrm{pers}}(t + \tau)
  \;=\; \mathrm{v}(t),
\end{equation}
where $\mathrm{v} = (u, v)$ are the 10-m zonal and meridional wind
components. The same definition is applied independently for
$\tau = 0, 1, \ldots, 180$\,min.
Because each scatterometer overpass covers only a portion of the
study domain, the persistence forecast inherits the spatial mask
$m_{t}$ of the initialisation observation. When scored
against the target observation at $t + \tau$, which has its own
spatial mask $m_{t+\tau}$, the comparison is restricted to
the intersection $m_{t} \cap m_{t+\tau}$. We apply this evaluation procedure also to WindCastNet and to the
regridded MEPS forecasts, so that the three forecast approaches – WindCastNet, MEPS and persistence – are scored on
identical grid cell sets at each lead time and the resulting RMSE values
are directly comparable.
The persistence error tends to grow continuously with increasing $\tau$, reflecting the
decorrelation of the wind field over time. Note that the
irregular sampling of low-Earth-orbit scatterometers introduces an
additional, sampling-driven source of variability that is specific to this evaluation setup: when a validation overpass happens to fall
close in time to the most recent input observation, the effective
time gap is small and the persistence error is artificially low,
whereas larger gaps produce correspondingly larger errors.

Because scatterometer overpasses are not synchronised with fixed forecast cycle times, 
an exact temporal gap of $\tau$ is rarely available. We therefore admit a tolerance of $\pm$ 15 minutes around the nominal lead time. For example, the $+1$~h persistence error is computed from pairs of observations that are 45 to 75 minutes apart. This range is tolerable also due to the spatial resolution of the scatterometer-derived wind field data: at their 25\,km grid spacing, the wind field hardly evolves on time scales of 15 minutes.

\section{Results and Discussion} \label{sec:performances}

\begin{figure}[pos=H]
    \centering
    \includegraphics[width=1.0\linewidth]{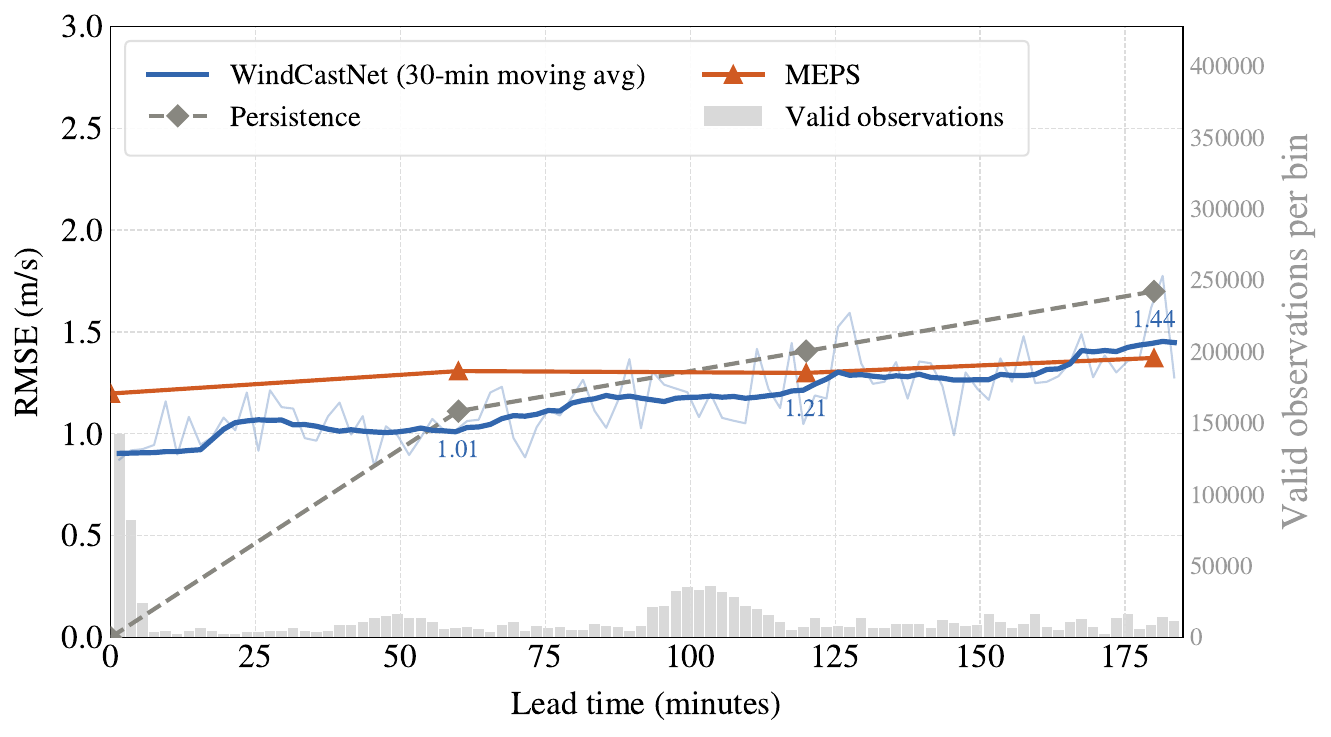}
    \caption{RMSE of 10-m wind speed as a function of forecast lead time for WindCastNet (blue), the persistence baseline (grey) and MEPS (orange), evaluated against scatterometer observations over the test set.}
    \label{fig:performances}
\end{figure}

\begin{figure}[pos=H]
    \centering
    \includegraphics[width=1\linewidth]{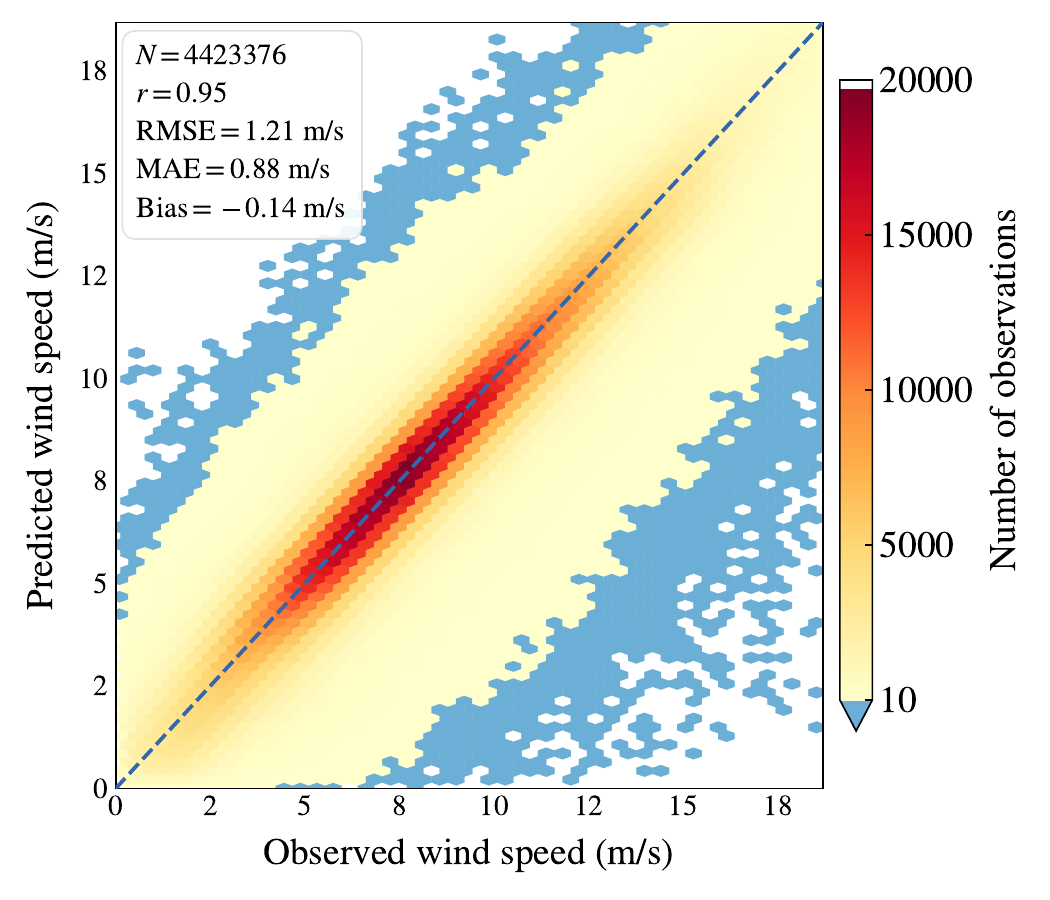}
    \caption{Two-dimensional histogram of WindCastNet predicted versus observed 10-m wind speed over the test set, aggregated across all the lead times up to 180 min.
    Blue hexagons mark bins with fewer than 10 occurrences, highlighting the tails of the distribution.
    }

    \label{fig:hexplot}
\end{figure}

 As shown in Fig. \ref{fig:performances} and Table \ref{table:performances}, forecast errors of all compared models grow with lead time. Compared to HARMONIE MEPS, WindCastNet reduces the RMSE of 10-m wind forecasts for lead times of one and two hours by 23\% and 7\%, respectively. WindCastNet outperforms HARMONIE MEPS 10-m wind forecasts for lead times of up to 2.5 hours, as shown in Fig. \ref{fig:performances}.
 Since MEPS provides forecasts at hourly time intervals, its RMSE is evaluated at the same temporal resolution. WindCastNet generates forecasts at various minute-level timestamps at irregular sampling intervals determined by the satellites' orbits, so samples falling precisely at 60, 120, and 180 minutes intervals between observations are sparse. The RMSE values of WindCastNet in Fig. \ref{fig:performances} therefore represent averages over a $\pm 15$-minute window centered on each nominal lead time. 
 The grey histogram at the bottom shows the amount of grid cells observed at each timestamp, with the number of observations aggregated into bins of 5 minutes. 
 Fig. \ref{fig:hexplot} presents the overall performance of WindCastNet forecasts against observed 10-m wind speeds based on more than 4.4 million observations across the study domain on the test set for forecast lead times of up to 180 minutes. Predictions and observations align closely around the $1{:}1$ line, so the predicted and observed offshore wind speeds are in close agreement, with a Pearson correlation coefficient of approximately $r = 0.95$, a bias of $-0.14$\,m/s, and a mean absolute error (MAE) of $0.88$\,m/s.

\begin{table}[pos=H]
\centering
\footnotesize   
\begin{tabular}{|l|l|c|c|c|}
\hline
\textbf{Lead Time} & \textbf{Metric} & \multicolumn{3}{c|}{\textbf{Wind Speed (m/s)}} \\
\cline{3-5}
 & & WindCastNet & MEPS & Pers. \\
\hline
\multirow{2}{*}{$+1\,\mathrm{h}$} & RMSE & $1.01$ & $1.31$ & $1.11$ \\
& Bias & $-0.12$ & $-0.20$ & $-0.03$ \\
\hline
\multirow{2}{*}{$+2\,\mathrm{h}$} & RMSE & $1.21$ & $1.30$ & $1.41$ \\
& Bias & $-0.17$ & $-0.18$ & $-0.02$ \\
\hline
\multirow{2}{*}{$+3\,\mathrm{h}$} & RMSE & $1.44$ & $1.37$ & $1.70$ \\
 & Bias & $-0.24$ & $-0.10$ & $-0.02$ \\
\hline
\end{tabular}
\caption{Comparison of wind speed RMSE and bias from WindCastNet, MEPS and persistence across lead times, evaluated on the test set.}
\label{table:performances}
\end{table}

\begin{figure*}[pos=H]
    \centering
    \includegraphics[width=1\linewidth]{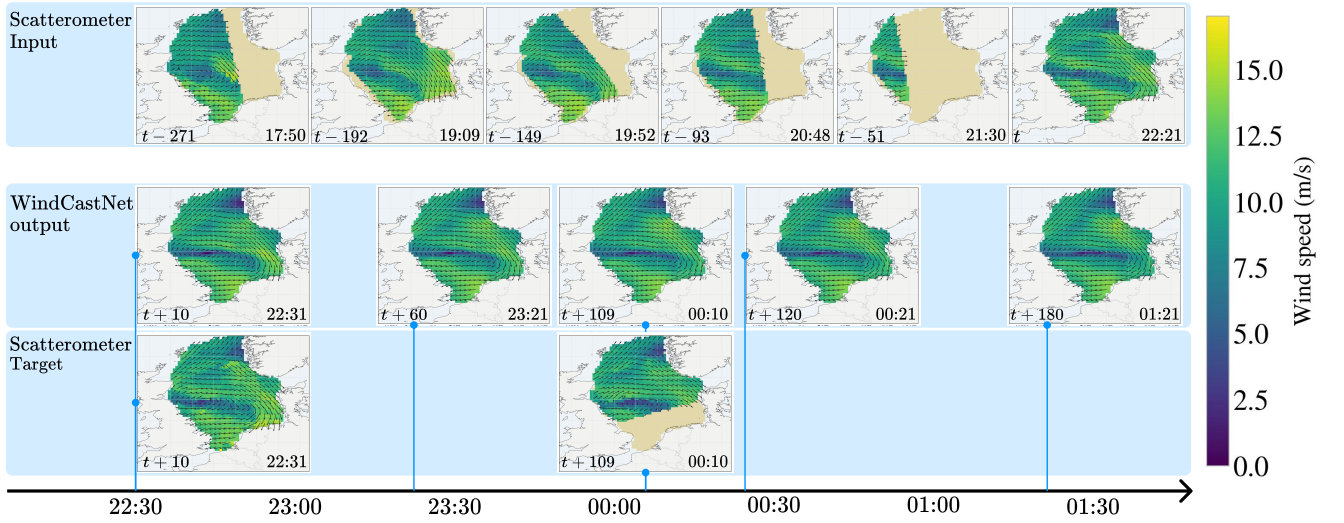}
    \caption{Comparison of near-surface wind forecasts from WindCastNet and scatterometer observations (input and target). Colour indicate wind speed, while vectors indicate wind speed and direction. Output and target columns are aligned by valid time along the axis below. WindCastNet output is shown both at target lead time (10, 109 minutes) and at fixed lead times (60, 120, 180 minutes). $t$ denotes the timestamp of the last observed input (22:21). }
    \label{fig:mode_output}
\end{figure*}

To further illustrate the performance of WindCastNet, a specific example forecast is shown in Fig. \ref{fig:mode_output}, comparing near-surface wind field forecasts from WindCastNet against actual scatterometer observations. The figure contains both the model's output at target lead times (10, 109 minutes) and at fixed lead times (60, 120, 180 minutes).
We also provide an example of the WindCastNet and MEPS output compared to the scatterometer observations in appendix, Fig \ref{fig:model_output_comparison}. Although WindCastNet produces per-minute forecasts, we show the forecasts with a 30-minute frequency. Since the last WindCastNet input occurs at time 17:56, its forecast at 1\,h lead time is shown at 18:56 and compared with the MEPS forecast at time 19:00. The same applies for 2\,h and 3\,h lead times.

\subsection{Spatial distribution of forecast errors}
\label{sec:spatialerror}

The spatial distributions of forecast errors and biases are shown in Figs. \ref{fig:rmse_comparison} and \ref{fig:bias_comparison} for lead times of up to three hours. WindCastNet forecast errors are lower than MEPS errors at lead times of between two and three hours, with significant differences across the study domain. Forecast errors increase with lead time, growing faster in WindCastNet. Across all lead times, WindCastNet errors are lowest in the central North Sea and increase towards the coastlines, which coincide with the boundaries of the study domain. This behaviour is more pronounced at longer lead times. 
In WindCastNet, a possible reason is that near-surface flow patterns originating outside the study domain - such as low-pressure systems over the Atlantic - cannot be anticipated by WindCastNet, as information from those regions is not used during either training or inference. The convolutional architecture is limited to the fixed spatial extent of the input grid and therefore cannot exploit information from outside the computational domain. Consequently, approaching weather systems are only observed once they enter the domain. 
This current limitation of WindCastNet can be addressed in future versions of the model, motivating further research on domain extension and boundary condition modeling.

Forecast biases are lower in WindCastNet compared to MEPS, even though slowly increasing with forecast lead time. MEPS significantly underestimates near-surface wind speeds along the North Sea coastlines, in the Southern Bight and across much of the Norwegian sector, by approximately 0.5–1.5 m/s at all lead times considered, as shown in Fig. \ref{fig:bias_comparison}. 
A similar coastal pattern can be seen in the RMSE of Fig. \ref{fig:rmse_comparison} where MEPS errors are lowest over the central North Sea and increase in a narrow band along the coastline at every lead time. MEPS forecasts also exhibit a moderate but systematic overestimation of wind speed in the western and central North Sea, off the east coast of Scotland and northern England. Biases in WindCastNet 10-m wind speed tend to be smaller than the MEPS biases, showing a moderate negative underestimation in several North Sea areas with increasing lead time.

A possible explanation of the significant MEPS biases near coastlines (Fig. \ref{fig:bias_comparison}) may be related to the fact that wind forecasts were found to be more skilful over ocean than over land in an intercomparison of high-resolution Arctic models that included HARMONIE AROME-Arctic \citep{kolzow}, which shares much of its formulation with MEPS.
MEPS's coastal errors appear consistent with a deficiency in how the model represents the land-sea transition. 
In principle, it is also conceivable that a minor part of the MEPS biases might be related to differences in land versus ocean wind statistics in that a $25$\,km grid cell blending native land and ocean grid points could produce minor biases during upscaling. However, since we evaluate the RMSE and biases against scatterometer measurements that do not contain any land pixels, 
our evaluation of MEPS forecasts
is restricted to ocean-only grid points also near the coast, which rules out this potential explanation for the biases.

\begin{landscape}
\vspace{-10cm}
\begin{figure}[H]
    \centering

    \begin{subfigure}{\linewidth}
        \begin{adjustbox}{width=0.75\paperheight, center}
            \includegraphics{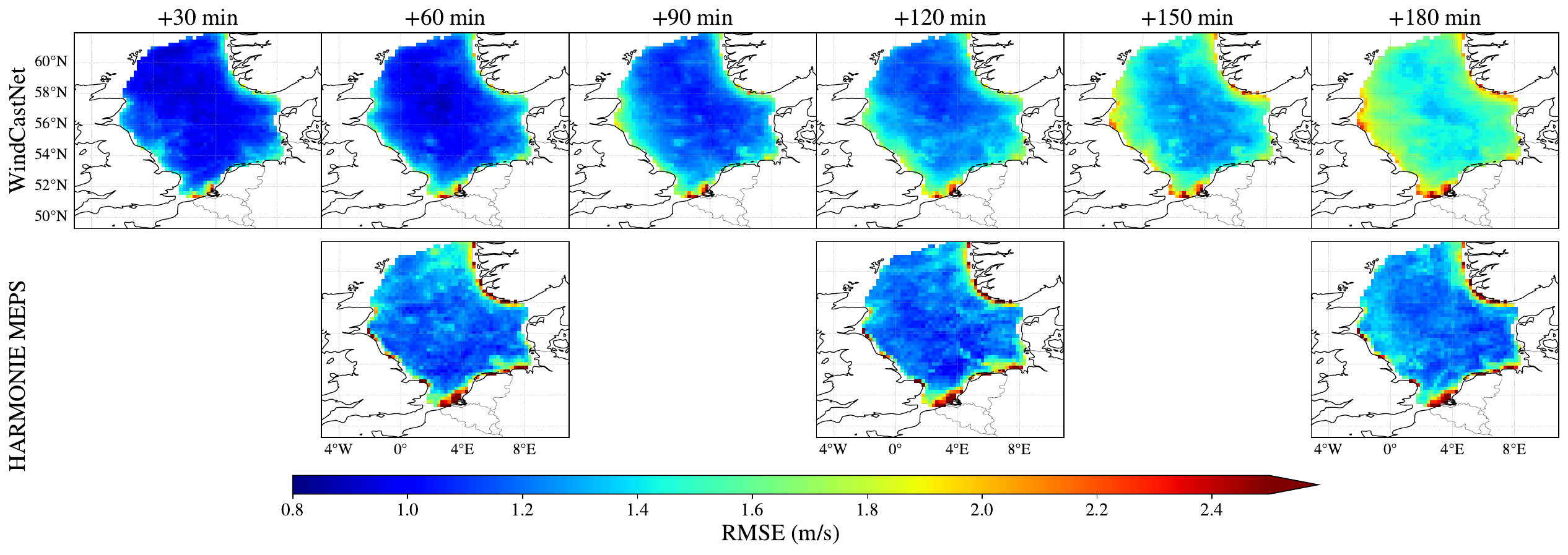}
        \end{adjustbox}
        \caption{Root Mean Square Error  
        (m/s) across the study domain.}
        \label{fig:rmse_comparison}
    \end{subfigure}

    \begin{subfigure}{\linewidth}
        \begin{adjustbox}{width=0.75\paperheight, center}
            \includegraphics{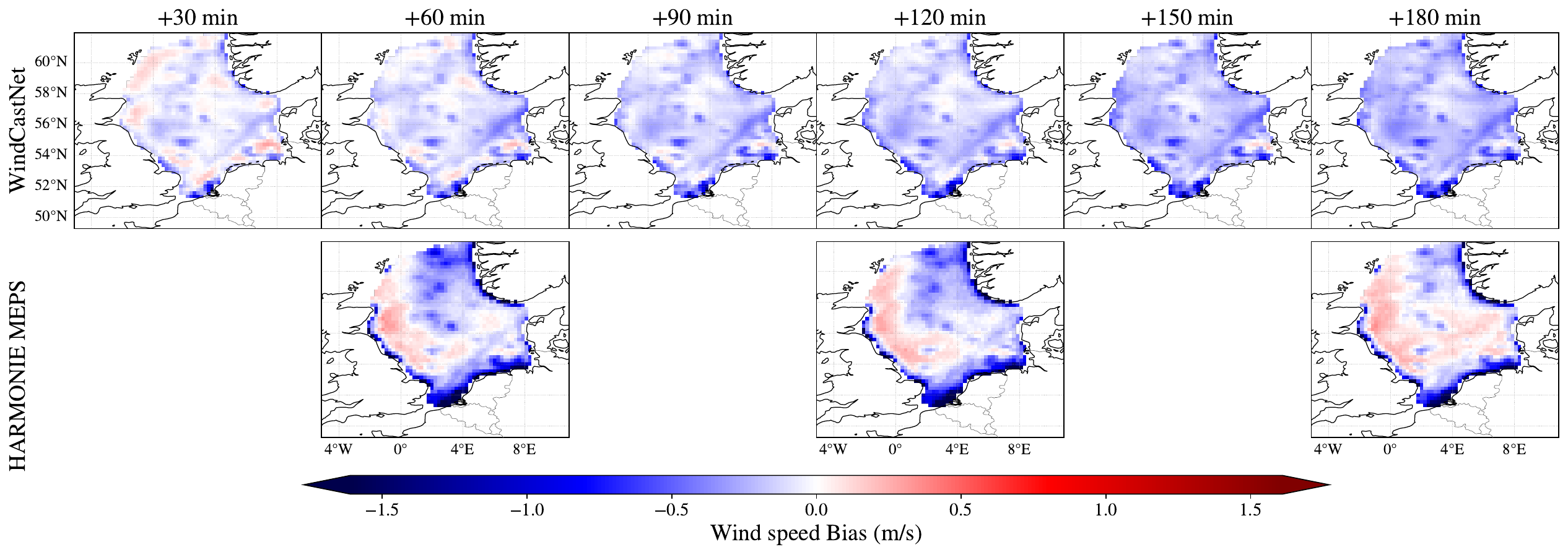}
        \end{adjustbox}
        \caption{Mean Bias Error
        (m/s) across the study domain.}
        \label{fig:bias_comparison}
    \end{subfigure}
    \caption{Spatial distribution of 10-m wind RMSE and Mean Bias Error.} 
\end{figure}

\end{landscape}

\subsection{Performance across wind speed regimes}
To further characterise the model, we investigate the WindCastNet forecast performance by wind speed regime. To this end, 
we split the test set into \textit{low} ($\mu_{ws} < 5$ m/s), \textit{medium} ($5 \leq \mu_{ws} < 10$ m/s) and \textit{high} wind speeds ($\mu_{ws} \geq 10$ m/s), where $\mu_{ws}$ is the average wind speed of each wind field observation (e.g. the ones shown in Fig. \ref{fig:scatterometer_measurement_1} and Fig. \ref{fig:scatterometer_measurement_2}. 
Table~\ref{tab:performance_by_regime} reports the WindCastNet performance by wind speed regime. WindCastNet forecast errors increase by lead time in all wind speed regimes, as expected. In addition, forecast errors tend to also increase with wind speed: higher wind speeds go along with moderately higher forecast errors.

This is as expected as stronger winds tend to be associated with more dynamic atmospheric conditions, in which both the magnitude and direction of the field can evolve rapidly.

\begin{table}[pos=H]
\centering
\small
\setlength{\tabcolsep}{4pt}
\begin{tabular}{l l *{3}{w{c}{1.4cm}}}
\toprule
& & \multicolumn{3}{c}{Wind speed RMSE [m/s]} \\
\cmidrule(lr){3-5}
Wind speed & Model & $1$\,h & $2$\,h & $3$\,h \\
\midrule
\multirow{2}{*}{Low}
  & WCN & 1.02 & 1.12 & 1.27 \\
  & MEPS        & 1.20 & 1.21 & 1.22 \\
\cmidrule(lr){1-5}
\multirow{2}{*}{Medium}
  & WCN & 1.00 & 1.16 & 1.33 \\
  & MEPS        & 1.28 & 1.27 & 1.33 \\
\cmidrule(lr){1-5}
\multirow{2}{*}{High}
  & WCN & 1.18 & 1.37 & 1.68 \\
  & MEPS        & 1.50 & 1.44 & 1.49 \\
\bottomrule
\end{tabular}
\caption{WindCastNet (WCN) and MEPS performance by lead time for different wind speed
regimes on the test set. The observed magnitudes are grouped into low ($\mu_{ws}<5$~m/s), medium (5~m/s $\leq \mu_{ws}<10$~m/s), and high ($\mu_{ws}\geq10$~m/s) 10-m wind speeds. 
}
\label{tab:performance_by_regime}
\end{table}

\subsection{Performance by wind field uniformity}
\label{par:homogeneity_index}
We further analyse the WindCastNet performance as a function of the \textit{spatial uniformity} of the near-surface wind field. A satellite-derived near-surface wind field is considered as mostly \textit{uniform} if the wind is predominantly unidirectional and as mostly \textit{non-uniform} if the wind direction exhibits strong spatial variability across the study domain, for example, in the case of low pressure systems.

Table~\ref{tab:performance_by_homogeneity} shows how WindCastNet's performance depends on the uniformity of the near-surface wind field. Forecast errors are larger in non-uniform wind conditions compared to more uniform wind conditions across the study domain. Moreover, forecast errors increase with lead time in both uniform and non-uniform wind fields, consistent with the findings presented in the previous sections.

To classify each test sample as \textit{uniform} or
\textit{non-uniform}, we quantify the directional coherence of the
observed wind field with a single scalar derived from directional statistics. Let
$\mathrm{v}_i = (u_i, v_i)$ denote the wind vector at the $i$-th valid grid
cell of a target frame, and let $\hat{\mathrm{v}}_i =
\mathrm{v}_i / \|\mathrm{v}_i\|$ be the corresponding unit vector,
defined wherever $\|\mathrm{v}_i\| > 0$. We define the
magnitude of the average unit wind-direction vector $R$ as

\begin{equation}
R \;=\; \left\| \frac{1}{N} \sum_{i=1}^{N} \hat{\mathrm{v}}_i \right\|
\;=\; \sqrt{\bar{u}^2 + \bar{v}^2},
\label{eq:R}
\end{equation}

where $\bar{u} = \tfrac{1}{N}\sum_i \hat{u}_i$ and
$\bar{v} = \tfrac{1}{N}\sum_i \hat{v}_i$ are the averages of the unit
vector components over the $N$ valid grid cells and where, by construction,
$R \in [0, 1]$. $R = 1$ corresponds to a perfectly uniform wind field in which
all wind vectors point in the same direction, while significantly smaller $R$ values correspond to less uniform flows. 
For each target frame, we compute $R$ over the valid cells. The sample-level coherence is then defined as the mean of $R$ across the 
three wind field forecasts corresponding to lead times of +1\,h, +2\,h, +3\,h. We label each sample in accordance with
\begin{equation}
\text{sample label} =
\begin{cases}
\text{uniform}   & \text{if } R \geq R_\text{thr}, \\
\text{non-uniform} & \text{if } R < R_\text{thr},
\end{cases}
\label{eq:label_homogeneity}
\end{equation}
where the threshold $R_\text{thr} = 0.85$ is chosen empirically to separate
more uniform flow
from non-uniform conditions.
The resulting partition of the test set yields
1124,  1080, and 689 uniform samples and 505, 449, and 307
non-uniform samples at the $+1$\,h, $+2$\,h and $+3$\,h lead times,
respectively, corresponding to a uniform fraction of approximately
$70\%$ over the full test set.

\begin{table}[pos=H]
\centering
\small
\setlength{\tabcolsep}{4pt}
\begin{tabular}{l l *{3}{w{c}{1.4cm}}}
\toprule
& & \multicolumn{3}{c}{Wind speed RMSE [m/s]} \\
\cmidrule(lr){3-5}
Regime & Model & $1$\,h & $2$\,h & $3$\,h \\
\midrule
\multirow{2}{*}{Uniform}
  & WCN & 1.01 & 1.15 & 1.35 \\
  & MEPS        & 1.29 & 1.25 & 1.32 \\
\cmidrule(lr){1-5}
\multirow{2}{*}{Non-uniform}
  & WCN & 1.18 & 1.40 & 1.66 \\
  & MEPS        & 1.39 & 1.38 & 1.41 \\
\bottomrule
\end{tabular}
\caption{WindCastNet (WCN) and MEPS performance by lead time and by spatial uniformity of the observed wind fields of the test set. Uniform
samples correspond to spatially uniform wind regimes (e.g.\ broad
unidirectional flow), while non-uniform samples correspond to wind fields
exhibiting stronger spatial variability, for example, associated with cyclones or fronts. The reported metrics are the wind speed RMSE of
$\mu_{ws}$.}
\label{tab:performance_by_homogeneity}
\end{table}
\subsection{Model design analysis}
\label{sec:ablation}

To quantify the model contributions to its overall performance, we conduct an ablation study in which one component at a
time is removed from the full model. The full model conditions its single output
frame on the requested lead time $\deltatime$ through three components:
a Time2Vec embedding of $\deltatime$, a FiLM modulation of the top-layer
hidden state, and a convolutional decoder that injects the embedding as additional
channels before the final partial convolution. We ablate each component in isolation:
replacing Time2Vec with a plain linear embedding of $\deltatime$ (removing the
periodic terms), disabling the FiLM modulation, and disabling the decoder. We
additionally evaluate a no-$\deltatime$ reference in which both FiLM and the
decoder are disabled, so that $\deltatime$ no longer reaches the network and the
model is forced to predict at all horizons from the encoder state alone. The
partial-convolution masking is intrinsic to the recurrent backbone and is
therefore not ablated. All variants are trained on the same data split, with the
same optimiser configuration and the same number of epochs as the full model, and
are evaluated on the test set against scatterometer observations, with the wind-speed RMSE pooled overall and grouped into the $\deltatime$ ranges of Table~\ref{tab:ablation}.

\begin{figure}[pos=H]
    \centering
    \includegraphics[width=1\linewidth]{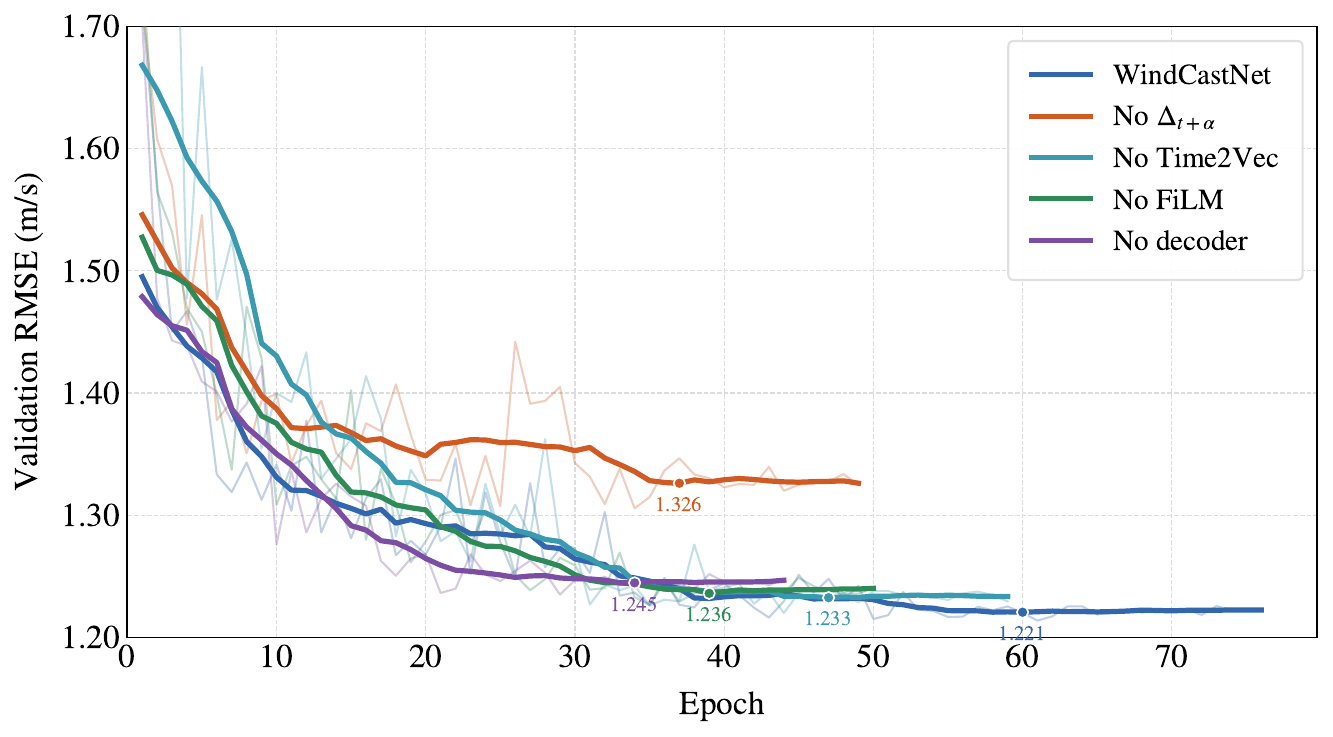}
    \caption{RMSE per epoch during training. The thin curves show the RMSE at each training epoch, whereas the thick curves show the corresponding moving average. The full WindCastNet model outperforms a WindCastNet architecture lacking the components indicated in the legend.}
    \label{fig:ablation_plot}
\end{figure}

Table \ref{tab:ablation} and Fig. \ref{fig:ablation_plot} report the wind-speed RMSE of each model variant. The lead-time
conditioning as a whole is the dominant contribution: removing it entirely (the
no-$\deltatime$ reference) raises the overall RMSE from $1.21$\,m/s to
$1.31$\,m/s ($+7.57$\%). Crucially, this degradation is strongly concentrated at
the shortest horizons. In the $<30$\,min bin of Table \ref{tab:ablation}, the RMSE rises from $1.14$ to
$1.30$\,m/s ($+14.4$\%) and in the $30$--$60$\,min bin from $1.11$ to
$1.22$\,m/s ($+10.3$\%), whereas beyond $120$\,min the gap shrinks to only
$+2.8$\% from 1.38 to 1.42\,m/s. This monotonic decay confirms that the
model genuinely exploits the requested lead time: lead-time conditioning is
most relevant when the field is still strongly correlated with the most recent
observation, while at longer $\deltatime$ the forecast tends to become less skilful. 

Among the remaining components,
the decoder contributes most ($+1.75$\% overall when removed), followed by FiLM ($+1.06$\%) and Time2Vec ($+0.51$\%). Replacing Time2Vec with a plain linear embedding of $\deltatime$ leaves the
performance almost unchanged, indicating that the periodic terms add only modestly beyond the linear lead-time signal. 
\begin{table}[pos=H]
\centering
\footnotesize
{\setlength{\tabcolsep}{4pt}
\begin{tabular}{lcccccc}
\toprule
Configuration & Overall & \multicolumn{5}{c}{RMSE by $\deltatime$ range [m/s]} \\
\cmidrule(lr){3-7}
              & [m/s] & $<30$ & $30$-$60$ & $60$-$90$ & $90$-$120$ & $120$-$180$ \\
\midrule
WindCastNet                       & 1.21 & 1.13 & 1.11 & 1.23 & 1.28 & 1.38 \\
$-$ Time2Vec     & 1.22 & 1.17 & 1.11 & 1.24 & 1.28 & 1.37 \\
$-$ FiLM         & 1.23 & 1.16 & 1.12 & 1.25 & 1.29 & 1.38 \\
$-$ Decoder      & 1.23 & 1.17 & 1.13 & 1.25 & 1.29 & 1.40 \\
$-$ $\deltatime$ & 1.31 & 1.30 & 1.22 & 1.33 & 1.34 & 1.42 \\
\bottomrule
\end{tabular}}
\caption{Ablation study on the test set. Each row removes one
component of the lead-time conditioning. The last row disables both FiLM and the decoder, recovering a $\deltatime$-blind model that predicts at every horizon from
the encoder state alone. RMSE is the wind-speed error, grouped by $\deltatime$ ranges in minutes.}
\label{tab:ablation}
\end{table}

\subsection{Limitations and future research}
Finally, we discuss limitations of the proposed framework and highlight opportunities for future research. First, the proposed forecasting method does not directly predict hub-height winds. The magnitude of the vertical wind shear depends on atmospheric stability and sea-surface roughness~\cite{Badger}. Nevertheless, near-surface winds over the oceans are generally well correlated with hub-height winds, such that accurate nowcasts of 10-m wind still provide highly valuable information, for example, on the timing and spatial evolution of wind ramps affecting turbine operation.
While NWP models resolve winds at heights closer to turbine hub heights, their forecasts usually remain subject to systematic biases and uncertainties arising from imperfect representations of boundary-layer processes~\citep{Kalverla2018,Myers2024,Jonas_2024,complementaryerrorstructures}. Our results demonstrate that, at lead times of minutes to a few hours, observation-driven wind nowcasting can accurately capture offshore wind evolution without being limited by the latency associated with data assimilation and numerical model integration. Future work could combine scatterometer observations with additional meteorological measurements and hub-height observations, for example, from meteorological masts, wind lidars, or turbine-mounted anemometers, to develop observation-based estimates of hub-height wind speeds.

Moreover, because the training domain is limited to the North Sea, WindCastNet cannot anticipate weather systems that originate outside the study domain, such as cyclones over the North Atlantic or the Norwegian Sea. Forecasts of non-uniform wind fields, which are associated with such large-scale synoptic features, are therefore more strongly affected. This limitation is expected to be substantially reduced by extending the training and inference domains. Although both were restricted to the North Sea in this study for computational reasons, the framework itself imposes no such a constraints. Expanding the domain to include adjacent regions would allow the model to observe approaching weather systems earlier while simultaneously increasing the size and diversity of the training dataset. Both factors are expected to improve forecasts of complex wind patterns and increase overall forecast skill.

Finally, WindCastNet was trained on a comparatively limited dataset derived from a scatterometer constellation. Forecast skill could likely be further improved by incorporating additional years of observations, expanding the spatial coverage, or integrating complementary data sources, such as NWP fields, ground-based observations, or additional satellite observations. These extensions represent promising directions for future work to further extend forecast horizons.

\section{Conclusions}
\label{sec:conclusions}
We introduced WindCastNet, an observation-driven framework for satellite-based offshore wind nowcasting that predicts near-surface wind fields directly from spatiotemporally irregular scatterometer observations. To our knowledge, this is the first demonstration of regional wind field nowcasting  and the first nowcasting framework designed to learn directly from low-Earth-orbit satellite observations and their irregular sampling in both space and time. By explicitly encoding spatial observation availability and continuous lead times, WindCastNet produces forecasts at arbitrary lead times without requiring regularly sampled input data.

Unlike operational NWP systems, which primarily rely on assimilating observations into physics-based models, WindCastNet learns directly from observed wind fields. This observation-driven paradigm avoids data assimilation and model integration, enabling computationally efficient forecasts while inherently representing atmospheric flow structures present in the observations. Operational NWP models such as MEPS typically do not represent wind farms and their effects as momentum sinks and wake-generating features. Observation-driven forecasting, however, learns from wind observations that inherently include wind-farm wake effects and other wind-farm-induced flow modifications.

WindCastNet outperforms the HARMONIE-AROME MEPS model at short lead times, reducing wind-speed RMSE by 23\% and 7\% at 1 and 2 hours ahead, respectively, while maintaining competitive performance for almost three hours. 
Forecast accuracy of WindCastNet is highest over the central North Sea and tends to decrease near the domain boundaries. 
Ablation experiments further demonstrate that explicitly accounting for the spatial and temporal irregularity of satellite observations is essential for achieving these improvements. 

We also identify several opportunities for future research. Because WindCastNet was trained only over the North Sea, it cannot anticipate weather systems developing outside the study domain. Extending the training domain is likely to both improve forecasts of approaching synoptic systems and increase the diversity of atmospheric conditions available for training. In addition, the current 25\,km resolution limits the representation of small-scale wind features, and the model predicts near-surface rather than hub-height winds. Future work should therefore explore higher-resolution observations, larger and more diverse training datasets, and additional observational data sources. Probabilistic forecasting and hybrid approaches that combine observation-based nowcasting with numerical weather prediction should also be explored.

Overall, our results demonstrate that irregular satellite scatterometer observations contain sufficient predictive information to enable accurate regional offshore wind nowcasting. As offshore wind power capacity and demand for high-frequency marine weather information continue to grow, observation-driven forecasting offers a computationally efficient complement to numerical weather prediction at lead times up to a few hours. Beyond offshore wind forecasting, this framework opens new opportunities for marine weather forecasting, including applications such as tropical cyclone forecasting, maritime operations, and coastal hazard monitoring.

\section{Acknowledgements} 
We acknowledge funding from the Swiss National Science Foundation under grant 229366. The code used in this study is publicly available at \\
https://github.com/EnergyWeatherAI/WindCastNet.

\section{Appendix}
This appendix provides supplementary material supporting the analyses presented in the paper.

\subsection{Training, validation and test sets}
\label{sec:train_val_test}
\begin{figure}[pos=H]
    \centering
    \includegraphics[width=1\linewidth]{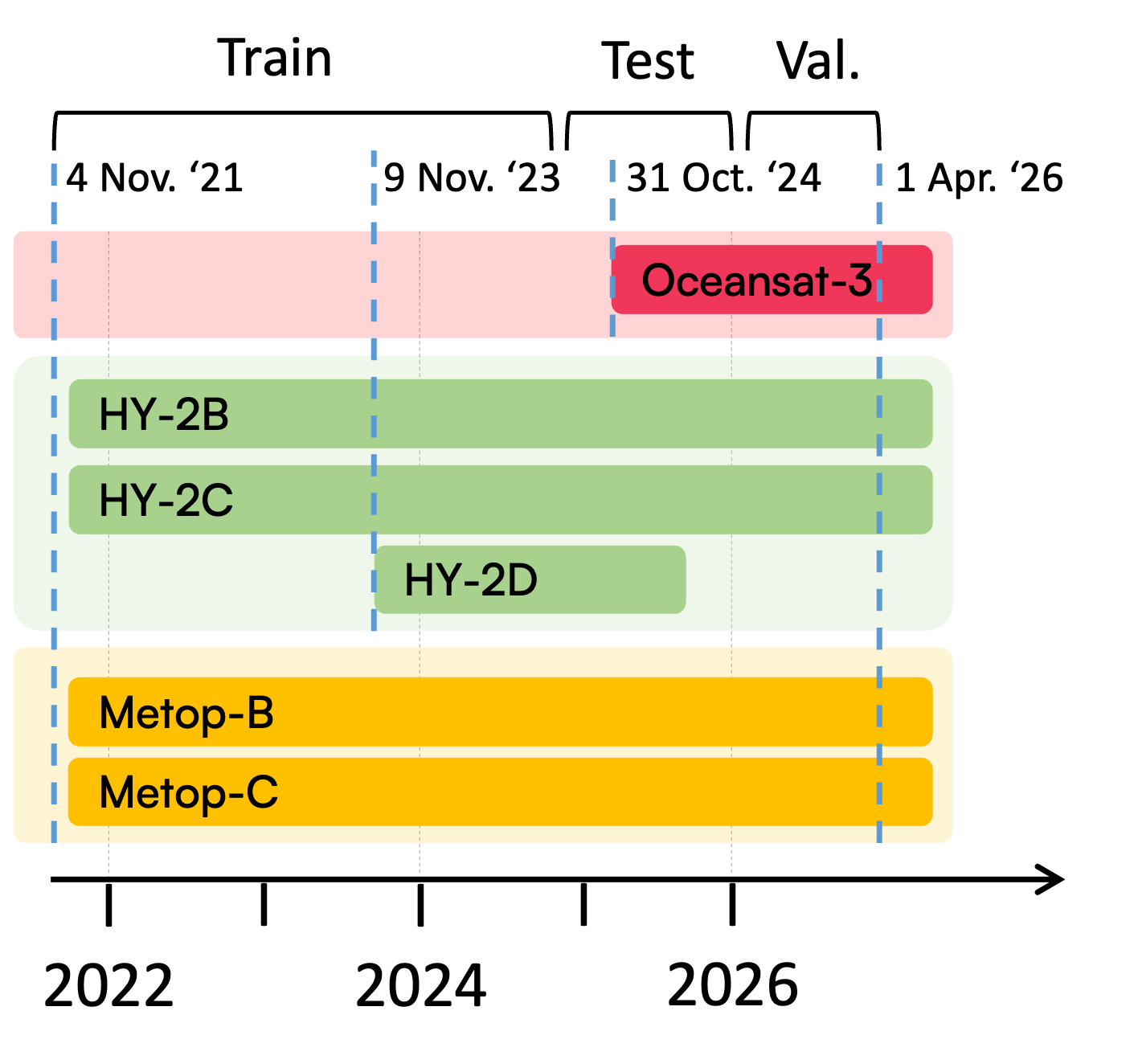}
    \caption{Satellite-based scatterometer observations: Training, validation and test sets.}
    \label{fig:dataset_timeline}
\end{figure}

The training, validation and
test periods employed in this study are the years
2021-2024, 2026 and 2025, respectively. In more detail, the training set includes all observations from HSCAT on HY-2B/2C and ASCAT MetOp-B/C between 4 November 2021 and 31 December 2024, as well as HY-2D from 9 November 2023 to 11 November 2024, and the OSCAT on Oceansat-3 from 31 October 2024 to December 2024, whereas the test set contains all observations from 1 January 2025 to 31 December 2025, and the validation set contains the Oceansat-3, HY-2B/C, MetOp-B/C from 1 January 2026 to 1 April 2026. Figure \ref{fig:dataset_timeline} shows the structure of the train, validation and test datasets based on the availability of the satellite data.

\begin{figure*}[pos=H]
    \label{fig:satellite_overpasses}
    \centering
    \begin{subfigure}{0.48\textwidth}
        \centering
        \includegraphics[width=1.2\linewidth]{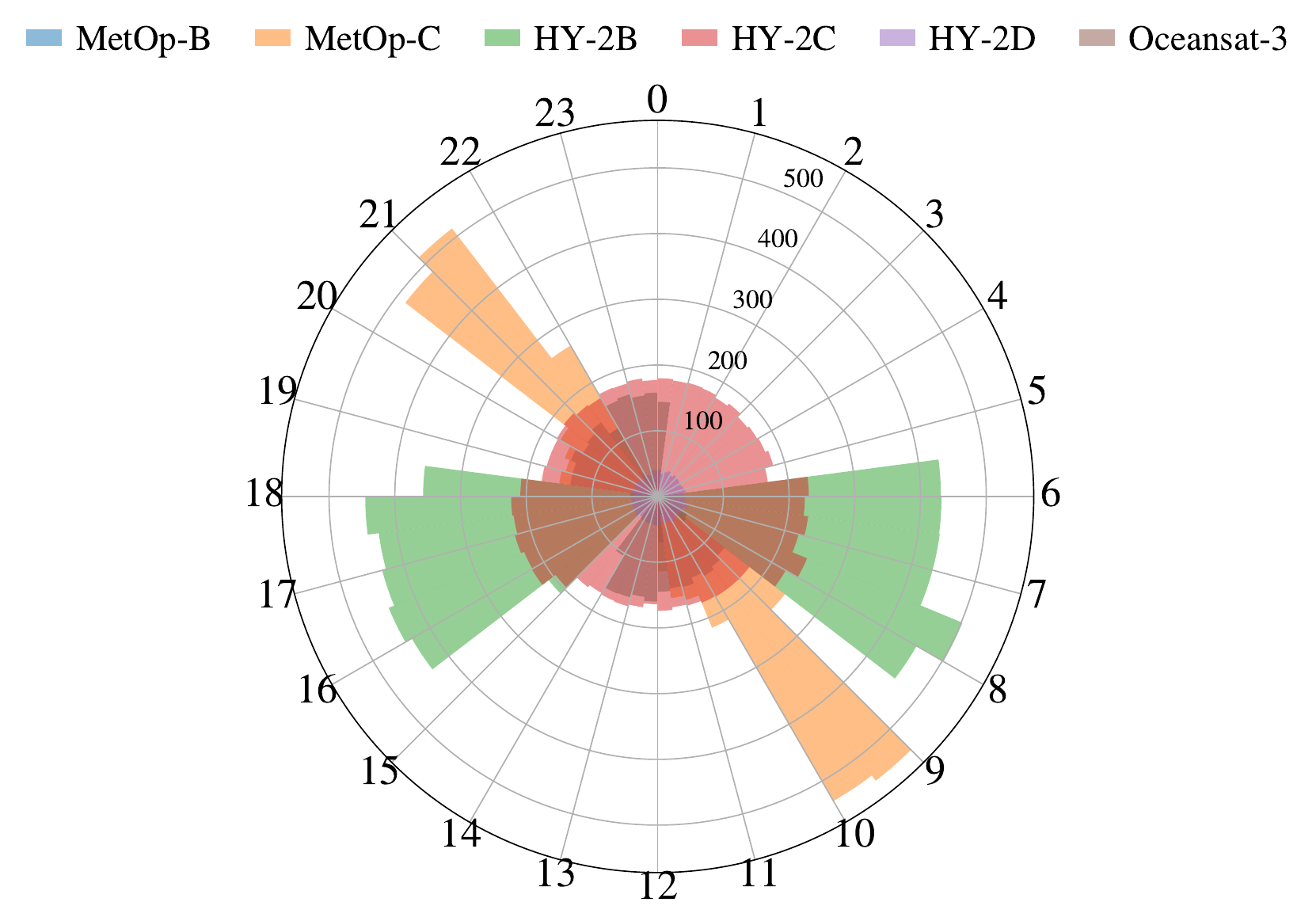}
        \caption{Overpass frequency over the study domain by satellite.}
        \label{fig:dataset_overpasses}
    \end{subfigure}
    \hfill
    \begin{subfigure}{0.48\textwidth}
        \centering
        \includegraphics[width=0.76\linewidth]{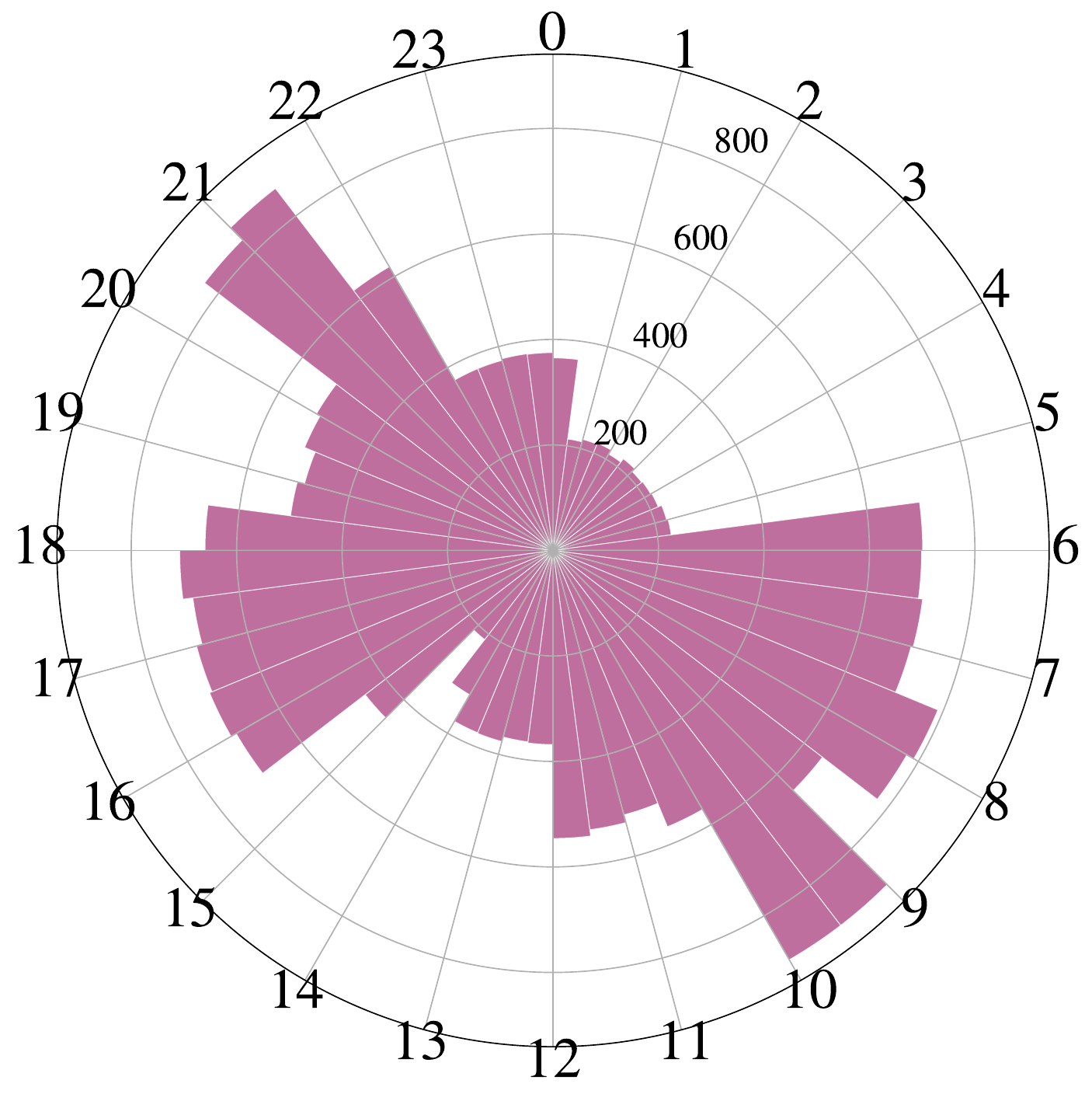}
        \caption{Overall distribution of the overpasses above the study domain.}
        \label{fig:overpasses_overall}
    \end{subfigure}
    \caption{Satellite overpass frequencies over the study domain by hour of the day in UTC.}
\end{figure*}

Since the various scatterometers of the satellite constellation were launched at different times over the past years, the daily overpass frequency above the study domain is not uniform across our dataset, ranging from as few as two to as many as 19 overpasses per day, depending on the amount of satellites providing observations operationally and their maintenance states. From November $4^{th}$ 2021 till November $9^{th}$ 2023, approximately 8 to 13 overpasses per day were available and included from satellites MetOp-B, MetOp-C, HY-2B and HY-2C. Since the release of HY-2D L2 wind observations on November $9^{th}$ 2023 till the availability of the Oceansat-3 data on October $31^{st}$ 2024, the amount of suitable overpasses per day reaches a peak of 17 observations per day. From October $31^{st}$ 2024 onwards, following the release of the OSCAT L2 wind data, the peak increases to 19 observations per day. 
Figure \ref{fig:dataset_timeline} gives an overview of the satellite coverage and corresponding L2 wind data availability over time.
As the sun-synchronous scatterometers overpass the study domain at approximately the same time each the day, we have a higher frequency of measurements in the intervals 5:30am-8:30am, 9:00am-10:00am, 3:30pm-6:00pm and 8:30pm-9:30pm UTC, as shown in Fig. \ref{fig:dataset_overpasses} and Fig. \ref{fig:overpasses_overall}.

\subsection{Optimal time window length}

\begin{figure}[pos=H]
    \centering
    \includegraphics[width=1\linewidth]{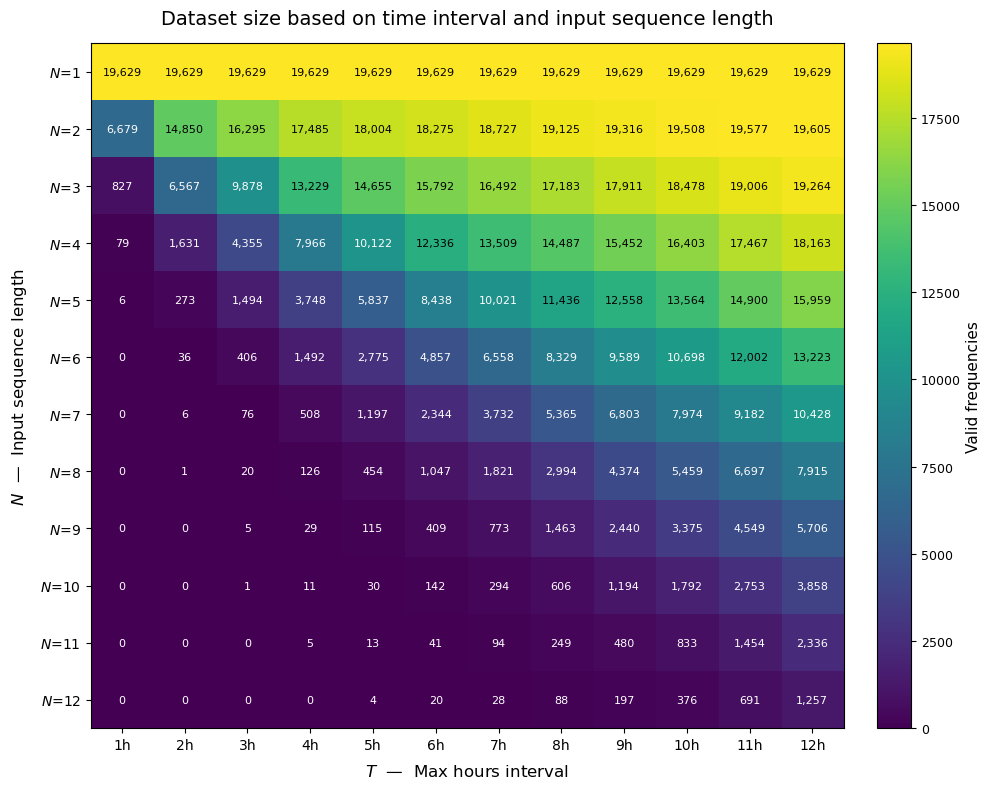}
    \caption{Dataset size as a function of $N$ and $T$.
    There are $13\,223$ cases with at least $6$ observations within $12$ hours.}
    \label{fig:tradeoff}
\end{figure}

Working with irregularly sampled wind observations requires identifying a suitable trade-off between the length $T$ of the time window during which observations are included in a forecast and the minimum number of observations $N$ available during that time window. These two hyperparameters determine the amount and the timeliness of the information available to the forecast model. 
A large time window provides more observations, but some of them may be temporally distant from the prediction time and therefore less representative of the current atmospheric state. Conversely, a short time window ensures more recent observations but may contain few or no samples, limiting the spatiotemporal context available to the model and potentially reducing forecast skill.
The choice of $T$ and $N$ therefore represents a balance between temporal relevance and contextual richness: 
The optimal configuration must provide enough observations to characterise the recent evolution of the wind field without relying excessively on measurements that are too old to remain informative. To identify the optimal combination of  $T$ and $N$, we performed a grid search over multiple combinations of the maximum time window $T$ and the number of observations $N$ (Fig. \ref{fig:tradeoff}). For multiple combinations, the model was trained and validated, and the validation RMSE was used as the selection criterion. The best performance was achieved with a configuration of $N=6$ observations within a maximum time window of $T=12$ hours.

\label{sec:dataset_tradeoff}

\subsection{Partial-convolutional LSTM cell}
\label{sec:partial_convlstm}
We provide a more detailed description of the partial-convolutional LSTM \cite{partialconvolutions, pconvlstm} in the following.
Given an input field $x$ with
binary mask (availability mask) $m$, a filter bank $W$ of kernel size $k$ and bias
$b$, the partial convolution is defined as
\begin{equation}
x' = W^{\top}\big(x \odot m\big)\,
\frac{k^{2}}{\,n_k(m)\,} + b,
\qquad
m' = \mathbb{1}\big[\,n_k(m) > 0\,\big],
\label{eq:pconv}
\end{equation}
where $n_k(m) = \mathbb{1}_k * m$ is the number of valid grid cells
in each $k \times k$ window, $\odot$ is the Hadamard product, and the output is
set to zero wherever $n_k(m) = 0$. The factor $k^{2}/n_k(m)$
rescales the response to be invariant to how many grid cells of the window were
actually observed, and the mask dilates by one convolution as it propagates.
In a partial-convolutional LSTM cell, the cell replaces the
input-to-state and state-to-state convolutions of a standard ConvLSTM with
mask-renormalised convolutions and carries a hidden mask alongside the hidden
state. Both the input $x_i$ and the previous hidden state
$h_{i-1}$ are masked,
\begin{equation}
x'_i = x_i \odot m_{x,i},
\qquad
h'_{i-1} = h_{i-1} \odot m_{h,i-1},
\label{eq:cell-mask}
\end{equation}
and the valid-grid cell counts in the input-to-state and state-to-state windows
(kernel sizes $k_x$ and $k_h$) give the renormalisation maps
\begin{equation}
S_x = \frac{k_x^{2}}{n_{k_x}(m_{x,i}) + \varepsilon}\,
\mathbb{1}\big[n_{k_x}(m_{x,i}) \ge 1\big],
\end{equation}
\begin{equation}
S_h = \frac{k_h^{2}}{n_{k_h}(m_{h,i-1}) + \varepsilon}\,
\mathbb{1}\big[n_{k_h}(m_{h,i-1}) \ge 1\big],
\label{eq:cell-scale}
\end{equation}
with $\varepsilon$ a small positive constant. The four gate pre-activations are produced
by a single pair of grouped convolutions, each renormalised by its valid
fraction and summed with a shared bias,
\begin{equation}
\big[i_i,\,f_i,\,g_i,\,o_i\big]
= \big(W_x * x'_i\big) \odot S_x
+ \big(W_h * h'_{i-1}\big) \odot S_h
+ b,
\label{eq:cell-gates}
\end{equation}
with $i_i, f_i, o_i = \sigma(\cdot)$ and
$g_i = \tanh(\cdot)$. The cell and hidden states follow the standard
ConvLSTM updates,
\begin{equation}
c_i = f_i \odot c_{i-1} + i_i \odot g_i,
\qquad
h_i = o_i \odot \tanh(c_i),
\label{eq:cell-state}
\end{equation}
and the hidden mask is propagated as the union of the valid input and previous
hidden positions within their windows,
\begin{equation}
m_{h,i} = \mathbb{1}\big[\,n_{k_x}(m_{x,i})
+ n_{k_h}(m_{h,i-1}) \ge 1\,\big].
\label{eq:cell-hmask}
\end{equation}
This formulation lets the cell condition its state transitions on the spatial
availability of the input and treats a hidden location as valid once it has
received information from either the observations or the recurrent state.
\subsection{Lead-time conditioning} \label{sec:leadtimeconditioning}
To produce forecasts at a requested lead time $\deltatime$, 
we keep the recurrent encoder
unchanged and condition only the output. The scalar lead time is first embedded
with Time2Vec, which combines one linear term with $d-1$ periodic terms,
\begin{equation}
\begin{split}
\big[\mathrm{t2v}(\deltatime)\big]_j =
\begin{cases}
\omega_j\,\deltatime + \varphi_j, & j = 0,\\
\sin\big(\omega_j\,\deltatime + \varphi_j\big), & 1 \le j \le d-1,
\end{cases}
\end{split}
\label{eq:t2v}
\end{equation}
with learnable frequencies $\omega_j$ and phases $\varphi_j$. A linear layer maps $e = \mathrm{t2v}(\deltatime)$ to per-channel parameters $(\gamma, \beta)$ that affine-modulate the
top-layer hidden state through a FiLM transformation (Eq. \ref{eq:FiLM}), and the embedding is also broadcast to a feature map $E$ and concatenated into a convolutional
decoder before the final partial convolution,
\begin{equation}
\gamma,\, \beta = \text{Linear}(e), 
\end{equation}
\vspace{-0.5cm}
\begin{equation}
h_{\mathrm{cond}} = \text{FiLM}\left(h^{\mathrm{top}}| \gamma, \, \beta  \right) = \left(1 + \gamma\right)\,h^{\mathrm{top}} + \beta,
\label{eq:FiLM}
\end{equation}
\vspace{-0.5cm}
\begin{equation}
h_{\mathrm{dec}} = \mathrm{ReLU}\Big(
\mathrm{Conv}\big([\,h_{\mathrm{cond}};\,E\,]\big)\Big),
\end{equation}
\vspace{-0.5cm}
\begin{equation}
\hat{x}(\deltatime) = \mathrm{PConv}\big(h_{\mathrm{dec}},
m^{\mathrm{top}}\big),
\label{eq:dt-head}
\end{equation}
where $\deltatime$ is the requested timestamp happening $\alpha$ minutes after the last input timestamp ($t$).  
Because $\deltatime$ enters only through the output path, the conditioning is
learned only when the model is trained over a range of horizons. Trained on a
single fixed horizon, the $\deltatime$ input is ignored.

\subsection{Loss metrics and evaluation}
\label{sec:lossmetrics}
The model is trained with a partial-convolution reconstruction loss adapted from
\cite{pconvlstm}, which combines a masked L1 loss 
term with
a total-variation (tv) smoothness term and treats target grid cells that were observed in
the input differently from those that were not. For a per-grid cell weight mask $w$,
we write the masked reconstruction error of the predicted field $\hat{y}$ against
the target $y$ (over the $u$ and $v$ components) as

\begin{equation}
    \ell(\hat{y}, y;\, w) =
    \frac{1}{B\,\Width\Height}
    \sum_{b,\,c,\,i,\,j}
    w_{b,i,j}\,\big|\hat{y}_{b,c,i,j} - y_{b,c,i,j}\big|,
    \label{eq:masked_l1}
\end{equation}

where the sum runs over the batch, the two wind components $c\in\{u,v\}$ and the
spatial grid, $B$ is the batch size, and $\Width,\Height$ are the grid width and
height. The channel sum is not normalised, so the two components are added rather
than averaged.

During training, we apply a \emph{missing-data} augmentation: each
already-incomplete input frame is further corrupted by dropping a fraction $p$ of
its valid grid cells at random (we use $p = 0.1$), and we record
\begin{equation}
    \Miobs = \mathbb{1}\Big[\textstyle\sum_{t=1}^{T} \tilde{m}_t > 0\Big],
\end{equation}
the set of grid cells observed at least once across the $T$ corrupted input masks
$\tilde{m}_i$. This forces the network to reconstruct grid cells it never saw in the input. The supervised target mask $m$, restricted to the central evaluation window
$\mathcal{R}$, is then split into the grid cells that were observed in the input and
those that were not:
\begin{equation}
    \mathcal{L}_{\text{iobs}} = \ell\big(\hat{y}, y;\, m \odot \Miobs\big),
    \,\,\,
    \mathcal{L}_{\text{mss}}  = \ell\big(\hat{y}, y;\, m \odot (1 - \Miobs)\big),
\end{equation}
A total-variation term regularises the composite field obtained by pasting the
known target grid cells onto the prediction,
$\tilde{y} = m \odot y + (1 - m) \odot \hat{y}$:
\begin{equation}
    \mathcal{L}_{\text{tv}} =
    \frac{1}{B\,\Width\Height}
    \sum_{\mathcal{R}}
    \Big( \big| \tilde{y}_{i+1,j} - \tilde{y}_{i,j} \big|
        + \big| \tilde{y}_{i,j+1} - \tilde{y}_{i,j} \big| \Big),
\end{equation}
evaluated over the central window $\mathcal{R}$. The full training objective is
\begin{equation}
    \mathcal{L}_{\text{train}} =
    \mathcal{L}_{\text{iobs}}
    + \lambda_m\, \mathcal{L}_{\text{mss}}
    + \lambda_{\text{tv}}\, \mathcal{L}_{\text{tv}},
\end{equation}
with $\lambda_m = 6$ and $\lambda_{\text{tv}} = 10^{-3}$ following the approach used in \cite{pconvlstm}. The larger weight $\lambda_m$ emphasises the never-observed grid cells that the model must
infer rather than copy from the input.
At validation and test time, the missing-data augmentation is disabled and
the reconstruction term is evaluated over all valid (central) target grid cells,
\begin{equation}
    \mathcal{L}_{\text{val}} =
    \ell(\hat{y}, y;\, m) + \lambda_{\text{tv}}\, \mathcal{L}_{\text{tv}}.
\end{equation}
As before, we use the root mean squared error of wind speed ($\windspeed$) on the
validation set as the model-selection criterion, retaining the checkpoint that
achieves the lowest value.

\begin{figure*}[pos=H]
    \centering
    \includegraphics[width=1\linewidth]{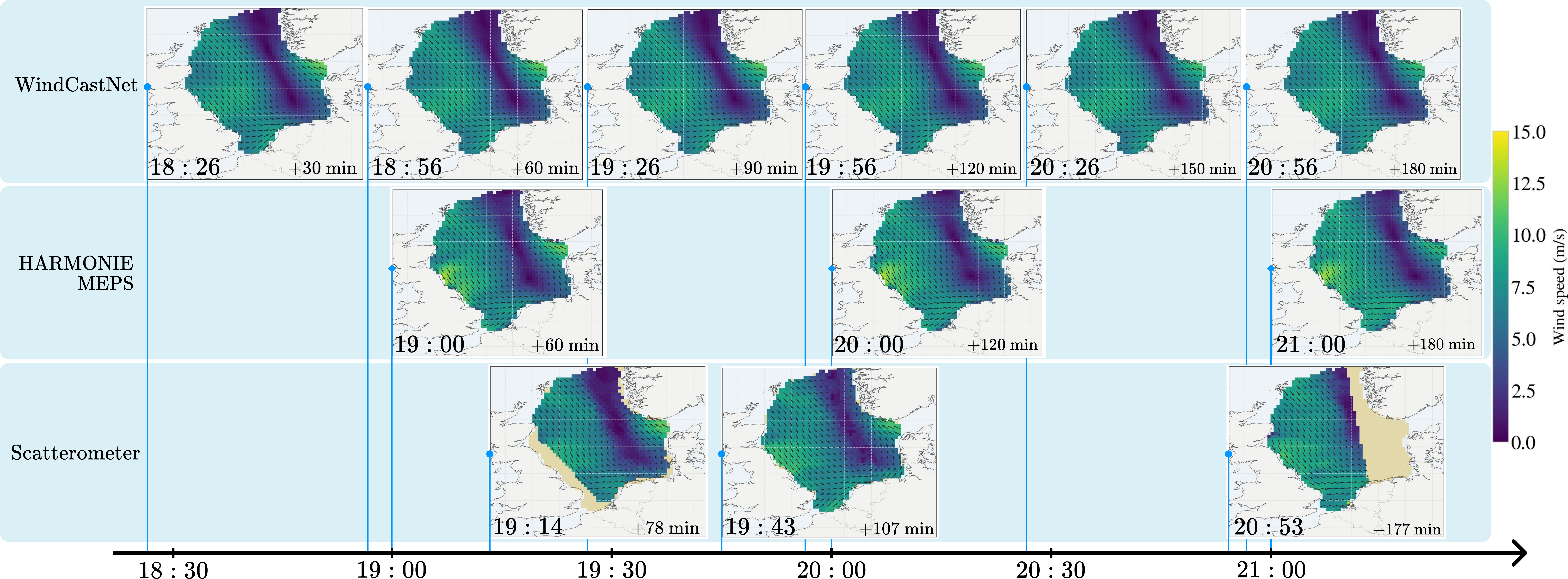}
    \caption{Comparison of near-surface wind forecasts from WindCastNet, HARMONIE-MEPS, and scatterometer observations
over the North Sea on 11 June 2025. Colour indicate wind speed, while vectors indicate wind speed and direction. Columns
are aligned by valid time along the axis below. WindCastNet output is shown every 30 minutes, whereas MEPS forecasts are
available hourly and scatterometer measurements only at irregular satellite-overpass times. $\Delta_t$ denotes the lead time relative to the initialisation time of WCN and MEPS. The last four input measurements used to generate the WindCastNet output in the
figure are shown in the input example (Fig. \ref{fig:input_data}).}
    \label{fig:model_output_comparison}
\end{figure*}

\bibliographystyle{cas-model2-names}
\bibliography{sample}   

\end{document}